\documentclass[final,12pt]{clear2025}

\title[Partial Cluster DAG]{Aligning Graphical and Functional Causal Abstractions}
\usepackage{times}
\usepackage{multirow}
\usepackage{hhline}
\usepackage{makecell}
\usepackage{rotating}

\usepackage{multicol}
\usepackage{float}
\usepackage{tikz}
\usetikzlibrary{cd}
\newtheorem{innercustomthm}{Theorem}
\newenvironment{customthm}[1]
  {\renewcommand\theinnercustomthm{#1}\innercustomthm}
  {\endinnercustomthm}
  
\newtheorem{innercustomlemma}{Lemma}
\newenvironment{customlemma}[1]
  {\renewcommand\theinnercustomlemma{#1}\innercustomlemma}
  {\endinnercustomlemma}

\newtheorem{innercustomcorollary}{Corollary}
\newenvironment{customcorollary}[1]
  {\renewcommand\theinnercustomcorollary{#1}\innercustomcorollary}
  {\endinnercustomcorollary}

\newcommand{\xdasharrow}[2][->]{
    \tikz[baseline=-\the\dimexpr\fontdimen22\textfont2\relax]{
        \node[anchor=south,font=\scriptsize, inner ysep=1.5pt,outer xsep=2.2pt](x){#2};
        \draw[shorten <=3.4pt,shorten >=3.4pt,dashed,#1](x.south west)--(x.south east);
    }
}

\newcommand{\ind}[0]{\perp\!\!\!\perp}
\newcommand{\nind}[0]{\not\!\perp\!\!\!\perp}
\newcommand{\model}[0]{\mathcal{M}}
\newcommand{\confarrow}[0]{\xdasharrow[<->]{\;\;\;\;\;}}

\newcommand{\vars}[0]{\mathbf{V}}
\newcommand{\clusts}[0]{\mathbf{C}}
\newcommand{\exos}[0]{\mathbf{U}}
\newcommand{\funcs}[0]{\mathcal{F}}
\newcommand{\dists}[1]{P_{#1}(\mathbf{U}_{#1})}
\newcommand{\relevant}[0]{\mathbf R}
\newcommand{\remain}[0]{\mathbf Q}

\newcommand{\lone}[0]{\mathcal{L}_1}
\newcommand{\ltwo}[0]{\mathcal{L}_2}
\newcommand{\lthree}[0]{\mathcal{L}_3}
\newcommand{\li}[0]{\mathcal{L}_i}
\newcommand{\domain}[0]{\mathcal{R}}

\newcommand{\alphabs}[0]{$\alpha$-abstraction}

\clearauthor{\Name{Willem Schooltink} \Email{Willem.schooltink@uib.no}\AND
 \Name{Fabio Massimo Zennaro} \Email{Fabio.zennaro@uib.no}\\
 \addr Department of Informatics, University of Bergen, Norway}

\begin{document}

\maketitle

\begin{abstract}%
  Causal abstractions allow us to relate causal models on different levels of granularity. To ensure that the models agree on cause and effect, frameworks for causal abstractions define notions of consistency. Two distinct methods for causal abstraction are common in the literature: (i) graphical abstractions, such as Cluster DAGs, which relate models on a structural level, and (ii) functional abstractions, like $\alpha$-abstractions, which relate models by maps between variables and their ranges. In this paper we will align the notions of graphical and functional consistency and show an equivalence between the class of Cluster DAGs, consistent $\alpha$-abstractions with the range of abstracted variables mapped bijectively, and constructive $\tau$-abstractions. Furthermore, we extend this alignment and the expressivity of graphical abstractions by introducing Partial Cluster DAGs. Our results provide a rigorous bridge between the functional and graphical frameworks and allow for adoption and transfer of results between them. 

\end{abstract}


\section{Introduction}
    Causality is a fundamental concept for understanding and predicting the behavior of complex systems. Uncovering the underlying causal mechanisms of a system is essential to make informed decisions and design more effective policies in critical fields such as medicine, economics, and politics.
The formalism of Structural Causal Models (SCMs) \citep{pearl2000models} provides us with a rigorous language to represent a causal system and reason about its behaviour not only in an observational regime ($\lone$), but also under interventions ($\ltwo$). 

Whatever causal system we consider, we always need to choose at which level of resolution we want to reason. For example, we may want to reason about voting behaviours either by defining a causal model on a person-by-person basis or by considering the causal dynamics at the level of districts. The two causal models represent the same system and can be connected by a relation of abstraction: individual voting behaviours can be aggregated into district voting patterns. While most causal algorithms  select a single among many possible levels of representation, switching between the levels can provide a richer understanding of a system; for instance, aggregating data at the individual and the district level may be valuable for designing successful advertisement strategies. 

We can express the relation between a low-level (or base) model and a high-level (or abstracted) model through a causal abstraction map. In order to dynamically switch between models, this abstraction map must guarantee the preservation of relations of cause and effect. We propose to distinguish two main ways to express abstractions and assess their consistency in the literature.
The first line of work on \emph{graphical abstractions} is based on the grouping or clustering of low-level variables, as proposed with the Cluster DAG (CDAG) approach \citep{anand2023cdag}. In these frameworks, we assess \emph{graphical consistency} in terms of the identifiability of causal queries across the base and the abstracted model; when all relevant interventional queries are correctly identifiable in both models, the abstraction is $\ltwo$\emph{-consistent}. 
The second line of work on \emph{functional abstractions} formalizes causal abstractions in terms of a map between variables and values in a low- and high-level model, as in the $\alpha$-abstraction \citep{rischel2020category} or the $\tau$-abstraction \citep{rubenstein2017causal} framework. In these frameworks, we evaluate \emph{functional consistency}  in terms of a discrepancy between the interventional distributions implied by the abstraction; in particular, whenever this discrepancy is zero for all relevant interventions, the abstraction is \emph{$\ltwo$-consistent}.  

As discussed in the related work, both frameworks have strong theoretical foundations and have provided the basis for relevant practical applications. To bring these frameworks and their methods together, in this paper we offer a formal bridging between graphical clustering and functional abstractions. Concretely, we work with CDAGs, which are the main graphical abstraction formalism in the literature, and with $\alpha$-abstractions, which offer the most explicit representation of a functional abstraction. To define a common ground of evaluation we first align the notion of functional $\ltwo$-consistency to graphical $\ltwo$-consistency; then, we show that the set of $\ltwo$-consistent CDAGs corresponds to a well-defined subset of simple $\alpha$-abstractions. Next, to increase the expressivity of graphical abstractions, we  introduce a natural extension of CDAGs, namely Partial CDAGs (PCDAGs) and prove that (i) PCDAGs describe a larger set of $\ltwo$-consistent $\alpha$-abstractions than CDAGs; and, (ii) under assumption of faithfulness, all bijective $\ltwo$-consistent $\alpha$-abstractions must be a PCDAG of the base model.  Last, we strengthen our contribution by rigorously showing an equivalence between two functional abstraction frameworks: the $\alpha$-abstraction and the constructive $\tau$-abstraction. This allows us to extend our connection between functional and graphical causal abstraction beyond the specific $\alpha$-abstraction framework. These results establish a firm connection between the different forms of abstraction proposed in the literature and their notions of consistency. From a theoretical point of view, our contributions allow for the transfer of proofs and properties between frameworks, while, practically, they suggest that PCDAGs may be a useful and grounded starting point for designing and validating consistent abstractions.

\paragraph{Related Work.}

Establishing the resolution of a SCM is a key design choice in the modeling of causal systems; the definition of variables and causal relations may be left to domain experts \citep{pearl2000models}, inferred from data through causal discovery or causal representation learning \citep{scholkopf2021crlsurvey}, or derived from pre-existing SCMs via graphical or functional causal abstraction.  

Seminal work on graphical models with clustered variables and their properties was published by \cite{PARVIAINEN2017110} and extended by \cite{wahl2024foundationsgroup}; \cite{anand2023cdag} introduced CDAGs as a causal inference tool for partially known causal models and proved results related to causal identifiability; CDAGs have also been adopted and interpreted as abstracted models in the context of learning abstractions using neural networks \citep{xia2024neuralabs}.

Functional causal abstraction comprise two main frameworks. An $\alpha$-abstraction \citep{rischel2020category, rischel2021compabstraction} defines two mappings between the variables and the values of two models; this framework has been used for learning abstractions \citep{zennaro2023jointly} and relate multi-armed bandits at multiple levels of abstraction \citep{zennaro2024causally}. Our work starts from the $\alpha$-abstraction framework as its formulation provides a more fine-grained understanding of an abstraction. On the other hand, a $\tau$-abstraction \citep{rubenstein2017causal,beckers2019abstracting,massidda2022causal} relies only on a single function between the values of two models; this framework has also been used to explain neural networks \citep{geiger2021causal}, learn abstractions \citep{felekis2023causal}, derive causal models targeted at encoding a chosen dynamics \citep{kekic2023targeted}, or generate surrogates of complex simulation models \citep{dyer2023interventionally}. We connect our results to a specific form of $\tau$-abstraction in the end of our work. Furthermore, \cite{massidda2024learningcausalabstractionslinear} have studied a linearized version of $\tau$-abstraction called $\mathbf T$-abstraction based on variable clustering; while the focus of their work was on abstraction for linear models and causal discovery, we study more generally the relation between clustering and functional abstraction.

\section{Preliminaries}
    In this section we first provide the notions of SCM and causal hierarchy; we then review graphical abstractions through the definitions of CDAGs and graphical consistency; and we conclude with functional abstractions through the presentation of the $\alpha$-framework and functional consistency. 

\paragraph{Notation.} We will denote a set of variables using bold uppercase $\mathbf V$, a specific variable using normal uppercase $V_i \in \mathbf V$, when necessary with an index subscript, and the value of that variable using lowercase $v_i$. We use $P(X)$ to denote a probability distribution; if necessary, we add a subscript $P_{\model}(X)$ to specify that the distribution is computed on model $\model$.  

\subsection{Causality}
\paragraph{Structural Causal Models.}
An SCM is a graphical and functional modelling tool that allows us to encode a causal system by (i) defining relations of cause and effects among variables, and (ii) specifying the behaviour of the variables via functions and probability distributions \citep{pearl2000models}.
    \begin{definition}[Structural Causal Model] \label{def:SCM} 
An SCM is a 4-tuple $\mathcal M: \langle \mathbf U, \mathbf V, \mathcal F, P(\mathbf U)\rangle$ where:
\begin{itemize}
    \item $\mathbf U$ is a set of exogenous variables each one with range $\domain(U)$. 
    \item $\mathbf V$ is a set of endogenous variables each one with range $\domain(V)$. 
    \item $\mathcal{F}$ is a collection of functions determining the value of the endogenous variables such that, for each $V \in \mathbf V$, there is a function $ f_V(\mathbf P_V, \mathbf U_V)$ with $\mathbf P_V\subseteq\vars\setminus V,\;\mathbf{U}_V\subseteq \mathbf U$. We say that $V_i \in \mathbf V$ is a \emph{direct cause} of $V_j \in \mathbf V$ if and only if $V_i$ is in the signature of the domain of the function $f_{V_j} \in \mathcal F$.
    \item $P(\mathbf{U})$ is the probability distribution over the exogenous variables $\mathbf U$. 
\end{itemize}
\end{definition}

In the following, we will make standard assumptions about our SCMs. We will restrict our attention to \emph{semi-Markovian} SCMs, that is, models without cyclic causal relations. It is immediate to show that such an SCM entails a Directed Acyclic Graph (DAG) $G = \langle \mathbf{V},\mathbf E\rangle$ with the set of vertices $\mathbf V$ given by the endogenous variables and the set of edges $\mathbf{E}$ given by the collection of  edges $V_i \rightarrow V_j$ if $V_i$ is a direct cause of $V_j$. The set $\mathbf P_{V_j}$ for a function $f_{V_j}$ represents then the graph-theoretical parents of ${V_j}$, and we will denote it as $Pa({V_j})$ from here on. Furthermore, if two variables $V_i,V_j\in\vars$ share an exogenous parent $U\in \exos$, we say that $V_i$ and $V_j$ are \emph{confounded}, implying they are not independent even without a causal edge between them; we denote confounding with a dashed bidirectional edge $V_i \confarrow V_j$ as a shorthand for $V_i \leftarrow U \rightarrow V_j$. Notice that, despite the bidirectional edge, the actual graph remains acyclic. 

Finally, we will also make the assumption of \emph{faithfulness}, that is two variables are graphically independent, or \emph{d-separated} \citep{geiger1990d}, if and only if they are distributionally independent. It is worth pointing out that, whereas an SCM $\mathcal{M}$ entails a unique DAG $G$, there normally are multiple SCMs $\mathcal{M}$ with the same DAG structure $G$.

Importantly, causality defines a way of interacting with SCMs via interventions. We consider hard interventions defined as follows:
\begin{definition}[Interventions]
    Let $\model=\langle \vars,\exos,\funcs,\dists{}\rangle$ be an SCM and $X\in\vars$ an endogenous variable, then an intervention $do(X=x)$ is an operation that replaces the function determining $X = f_X(Pa(X),U_X)$ with the assignment $X = x$.
\end{definition}
In analogy to conditioning, we will abbreviate the generic $do(X=x)$ as $do(X)$. Notice how an intervention $do(X)$ removes the direct causes of $X$; therefore it implies a new SCM with an underlying DAG with all the incoming edges into $X$ removed.  It is immediate to extend the definition of an intervention on a single variable $do(X)$ to multiple variables $do(\mathbf{X})$.

\paragraph{Causal Hierarchy.}
    A causal query on an SCM is a causal statement that can be located onto one of the three distinct layers of the Pearl's Causal Hierarchy (PCH) \citep{pearl2000models,Bareinboim2020hierarchy}: ($\lone$) observational queries with the form $P(Y|X)$ are statistical formulas that describe the system in its natural behaviour; ($\ltwo$) interventional queries with the form $P(Y|do(X))$ characterize the system under external manipulations; and, ($\lthree$), counterfactual queries study the system under hypothetical interventions that never took place.
These layers are rigid, as queries on a higher layer require more information to be answered and cannot automatically be reduced to a lower layer. In order to assess the consistency of abstractions, we will focus on the first two levels of this hierarchy. We will say that a causal query on the interventional layer $\ltwo$ is \emph{identifiable} if it can be reduced to statistical quantities belonging to the observational layer $\lone$. Under the assumption of faithfulness, \emph{Do-calculus} is a complete theory relying only on the graphical structure $G$ of $\model$ to decide whether a causal query is identifiable.

\subsection{Graphical Abstraction}

\paragraph{Cluster DAGs.}
The simplest way to relate two SCMs is through their underlying DAGs. CDAGs were originally introduced as graphical modelling tools allowing us to represent a system where information about the exact causal relationships among certain sets of variables were unknown \citep{anand2023cdag}. Nonetheless, CDAGs may also be easily used to express the process of reduction of information associated with abstraction. 
\begin{definition}[Cluster DAG]\label{def:CDAG}
     Let $G = \langle \mathbf{V},\mathbf E\rangle$ be a DAG and $\varphi:\mathbf{V} \rightarrow \mathbf{C}$ be a surjective function where $\mathbf C = \{\mathbf C_1, \dots, \mathbf C_k\}$. The function $\varphi$ induces a partition of $\mathbf{V}$. $G_{\mathbf C} = \langle \mathbf C, \mathbf E_{\mathbf C}\rangle $ is a cluster DAG (CDAG) of $G$ if and only if the set of edges $\mathbf E_{\mathbf C}$ abides by the following rules:
    \begin{enumerate}
        \item An edge $\mathbf C_i \rightarrow \mathbf C_j$ is in $\mathbf E_{\mathbf C}$ if there exists a $V_i \rightarrow V_j$ in $\mathbf E$ such that $V_i\in \mathbf C_i,\; V_j\in \mathbf C_j, \clusts_i \neq \clusts_j$.
        \item A bidirected (confounding) edge $\mathbf C_i \confarrow \mathbf C_j$ is in $\mathbf E_{\mathbf C}$ if there exists a $V_i \confarrow V_j$ in $\mathbf E$ such that $V_i \in \mathbf C_i, \; V_j \in \mathbf C_j, \clusts_i \neq \clusts_j$.
    \end{enumerate}
    Further, it is required that the graph $G_{\clusts}$ induced by $\varphi$ is acyclic.
\end{definition}

Given an SCM $\mathcal{M}$ with underlying DAG $G = \langle \mathbf{V},\mathbf E\rangle$, we can see the CDAG $G_{\mathbf C} = \langle \mathbf C, \mathbf E_{\mathbf C}\rangle $ as the structure of a graphically abstracted model $\model'$. Importantly, a CDAG $G_{\mathbf C}$ does not identify a single model $\model'$, but the collection of SCMs sharing the same structure.

\paragraph{Graphical Consistency.}
Although the building procedure of a CDAG guarantees the preservation of basic relations of cause and effect, we still want 
to establish whether a causal query can be consistently identified both in the original SCM $\model$ and in a model $\model'$ with the structure given by the derived CDAG $G_{\mathbf C}$. 
Since the SCM $\model'$ is not specified, we can investigate identifiability only by evaluating graphical properties.
We can express the relations of dependence and independence encoded graphically in a DAG through algebraic relation of equality and inequality among distributions; for instance, the structure $X \rightarrow Y$ implies, amongst others, relations such as $P(X\:|\:Y)\neq P(X)$ and $P(Y\:|\:X) = P(Y\:|\:do(X))$. 
We denote $\mathcal{G}(G)$ the set of all the algebraic constraints implied by the DAG $G$. We use these constraints to define graphical consistency:

\begin{definition}[Graphical Consistency]
Let $G$ be a DAG underlying $\model$ and $G_{\mathbf C}$ the CDAG underlying $\model'$. 
Let us define $\mathcal{G}(G_{\clusts}^{-1})$ be the set of all the constraints obtained by taking each constraint in $\mathcal{G}(G_{\clusts})$ and substituting each cluster $\clusts_i$ with its pre-image $\varphi^{-1}(\clusts_i)$. The two models $\model$ and $\model'$ are graphically consistent if $\mathcal{G}(G_{\clusts}^{-1}) \subseteq \mathcal{G}(G)$. 
\end{definition}

Graphical consistency means that all the equalities and inequalities expressed in the CDAG hold in the original SCM among the clustered variables; thus, identifiable causal queries on the CDAG are identifiable also on the original SCM. However, the converse does not necessarily hold: in the original SCM there may be additional equalities and inequalities among the variables within a cluster that are not expressible in the CDAG.

Graphical consistency may be restricted only to constraints pertaining to a chosen layer of the causal hierarchy. Considering again the structure $X \rightarrow Y$, the constraint $P(X\:|\:Y)\neq P(X)$ belongs to $\mathcal{L}_1$, while $P(Y\:|\:X) = P(Y\:|\:do(X))$ belongs to $\ltwo$. We denote $\mathcal{G}^{\mathcal{L}_i}(G)$ the set of all the constraints at layer $\mathcal{L}_i$ implied by the DAG $G$. We can express restricted forms of consistency as:

\begin{definition}[(Graphical) $\li$-Consistency] \label{def:Li-consistency}
Let $G$ be a DAG underlying $\model$ and $G_{\mathbf C}$ the CDAG underlying $\model'$. 
Let us define $\mathcal{G}^{\li}(G_{\mathbf C}^{-1})$ as the set of all the constraints obtained by taking each constraint in $\mathcal{G}^{\li}(G_{\mathbf C})$ and substituting each cluster $\clusts_i$ with its pre-image $\varphi^{-1}(\clusts_i)$. The two models $\model$ and $\model'$ are graphically consistent if $\mathcal{G}^{\li}(G_{\mathbf C}^{-1}) \subseteq \mathcal{G}^{\li}(G)$. 
\end{definition}

By construction, 
\sloppy a low-level model $\model$ and a high-level model $\model'$ with an underlying CDAG are always $\ltwo$-consistent; therefore, for a constraint $P_{\model'}(Y \vert X) = P_{\model'}(Y \vert do(X))$ in the high-level model, the constraint $P_{\model}(\varphi^{-1}(Y) \vert \varphi^{-1}(X)) = P_{\model}(\varphi^{-1}(Y) \vert do(\varphi^{-1}(X)))$ holds in the low-level model. In general, however, equalities \emph{across models}, such as $P_{\model'}(Y \vert X) = P_{\model}(\varphi^{-1}(Y) \vert do(\varphi^{-1}(X)))$, do not hold; however, given a CDAG, \citet{anand2023cdag} prove the existence of a high-level SCM $\model'$ with CDAG structure such that equalities across models hold (Theorem 7) and evaluations of do-calculus formulas on $\model'$ are equivalent to the evaluations on pre-images of the clusters in $\model$ (Theorem 3). 

\subsection{Functional Abstraction}

\paragraph{$\alpha$-Abstraction.}
Whereas a CDAG captures the graphical aspect of an abstraction, a functional abstraction is meant to express an abstraction in terms of a mapping between variables and values 
(see App. \ref{app:original-alpha} for the original definition with finite graphical models).

\begin{definition}[$\alpha$-abstraction]\label{def:alpha-abs} 
    \sloppy Given two SCMs $\mathcal M:\langle \vars_{\model},\exos_{\model},\funcs_{\model},\dists{\model}\rangle$ and $\mathcal M':\langle \vars_{\model'},\exos_{\model'},\funcs_{\model'},\dists{\model'}\rangle$, an abstraction $\boldsymbol{\alpha}: \mathcal M \rightarrow \mathcal M'$ is a 3-tuple $\langle \mathbf R, \varphi, \alpha_{V'} \rangle$ where:
    \begin{enumerate}
        \item $\mathbf R\subseteq \mathbf V_\mathcal M$ is a subset of relevant variables in $\model$.
        \item $\varphi: \mathbf R \rightarrow \mathbf V_{\model'}$ is a surjective map from the relevant variables to the variables in $\model'$.
        \item $\alpha_{V'}: \domain(\varphi^{-1}(V')) \rightarrow \domain(V')$, for each $V' \in \mathbf{V}_{\model'}$, is a surjective function from the range of the pre-image $\varphi^{-1}(V') \subseteq \vars_\model$ in $\mathcal M$ to the range of $V'$ in $\mathcal M'$.
    \end{enumerate}

\end{definition}

We will define the application of $\alpha_\vars$ to a distribution as the pushforward $\alpha_X\left[P(X)\right] = {\alpha_X}_{\#}(P)(X)$.

\paragraph{Functional consistency.}
Differently from the constructive definition of a CDAG, the declarative definition of an $\alpha$-abstraction does not implicitly preserve relations of cause and effect. 
A requirement of consistency over distributions is instead expressed in terms of commutativity of abstractions and $\li$ operations, such as conditioning ($\lone$) or intervening ($\ltwo$); we use the shorthand $P(Y\vert \li(X))$ for $P(Y\vert X)$ if $i=1$ or $P(Y\vert do(X))$ if $i=2$.

\begin{definition}[(Functional) $\li$-Consistency] Let $\boldsymbol{\alpha}: \model \rightarrow \model'$ be an abstraction. The abstraction $\boldsymbol{\alpha}$ is $\li$-consistent if, for all $ \mathbf X, \mathbf{Y} \subseteq \vars_{\model'}$, the following diagram commutes:
\begin{center}
    \begin{tikzcd}[column sep=3cm]
        \varphi^{-1}(\mathbf{X}) \arrow[r, "\li"]\arrow[d, "\alpha_\mathbf{X}"] & \varphi^{-1}(\mathbf{Y})\vert \li(\varphi^{-1}(\mathbf{X}))\arrow[d, "\alpha_\mathbf{Y}"]\\
        \mathbf{X} \arrow[r, "\li"] & \mathbf{Y}\vert \li(\mathbf{X})
    \end{tikzcd}.
\end{center}
that is:
\begin{equation}\label{eq:0abstractionerrror}
        P_{\model'}(\mathbf{Y}\:|\:\alpha_{\mathbf{X}}[\li(\varphi^{-1}(\mathbf{X}))]) = \alpha_{\mathbf{Y}}\left[P_{\model}(\varphi^{-1}(\mathbf{Y})\:|\: \li(\varphi^{-1}(\mathbf{X})))\right].
\end{equation}
\end{definition}

Thus, $\ltwo$-consistency means that the result of intervening and then abstracting must be identical to that of abstracting and then intervening \citep{rischel2020category}. In practical applications, however, we have to deal with noisy data and the need to drop information. As a result, it may be necessary to relax the strict requirement of commutativity and introduce an error measure that quantifies how different are the distributions on the two sides of Eq.\ref{eq:0abstractionerrror}:

\begin{definition}[$\li$--Abstraction error]
    Given an abstraction $\boldsymbol{\alpha}: \model \rightarrow \model'$ and a distance or divergence $D(p,q)$ between distributions $p$ and $q$, the $\li$-abstraction error is computed as:
    \begin{equation}\label{eqn:abstraction_error}
        e_{\li}(\boldsymbol{\alpha})=\max_{\mathbf X, \mathbf Y\subseteq \mathbf V_{\model'}} D(P_{\model'}(\mathbf Y\:|\:\alpha_{\mathbf{X}}[\li(\varphi^{-1}(\mathbf{X}))]), \alpha_{\mathbf{Y}}\left[P_{\model}(\varphi^{-1}(\mathbf Y)\:|\: \li(\varphi^{-1}(\mathbf X))\right]).
    \end{equation}
\end{definition}

Notice how the abstraction error is a worst-case measure of inconsistency: considering all possible operations in the abstracted model, it returns the error corresponding to the pair $\mathbf X,\mathbf Y \subseteq \vars_{\model'}$ that maximizes the discrepancy $D$. It is immediate to redefine functional $\li$-consistency as zero abstraction error $e_{\li}(\boldsymbol{\alpha}) = 0$.

\section{Aligning Graphical and Functional Consistency}

In order to relate graphical and functional abstractions, we need first of all to align the notions of graphical and functional consistency. Both $\li$-consistencies share the following implication:
\begin{lemma}\label{lem:l2-implies-l1-cons}
    $\ltwo$-consistency implies $\lone$-consistency. \emph{[Proof in App. \ref{proof-l2-implies-l1-cons}.]}
\end{lemma}

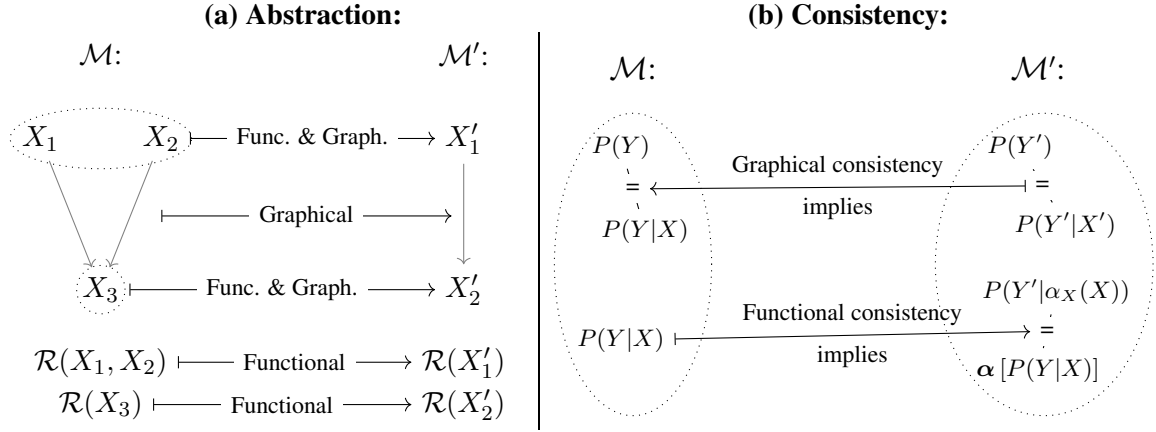
\begin{figure}
    \centering
    \textbf{(a) Abstraction:}\hspace{.3\textwidth}\textbf{(b) Consistency:}
    \begin{minipage}{0.46\textwidth}
    \centering
    \begin{tikzpicture}[scale=0.8]
    \node (M) at (0, 0.9) {$\model$:};
    \node (M) at (6, 0.9) {$\model'$:};
    \node (X1) at (1, -.5) {$X_2$};
    \node (X2) at (-1, -.5) {$X_1$};
    \node (X3) at (0, -3) {$X_3$};
    
    \node (Xp1) at (6, -.5) {$X'_1$};
    \node (Xp2) at (6, -3) {$X'_2$};

    \node (RX1) at (0, -4.2) {$\domain(X_1, X_2)$};
    \node (RXp1) at (6, -4.2) {$\domain(X'_1)$};
    \node (RX2) at (0, -4.9) {$\domain(X_3)$};
    \node (RXp2) at (6, -4.9) {$\domain(X'_2)$};
    
    \draw[->, gray] (X1) -- (X3) node[midway, above] (E1){};
    \draw[->, gray] (X2) -- (X3);
    
    \draw[->, gray] (Xp1) -- (Xp2) node[midway, above] (Ep1){};

    \draw[dotted] (0, -.5) ellipse (1.5cm and 0.5cm);
    \draw[dotted] (0,-3) ellipse (.4cm and .4cm);

    \draw[|->] (E1) ++ (.5cm,-.2cm) -- ([shift=({-0.2cm, -0.2cm})] Ep1.center) node[midway, centered, fill=white] {\footnotesize Graphical};
    
    \draw[|->] (RX1) -- (RXp1) node[midway, centered, fill=white] {\footnotesize Functional};
    \draw[|->] (RX2) -- (RXp2) node[midway, centered, fill=white] {\footnotesize Functional};

    \draw[|->] (X1) -- (Xp1) node[midway, centered, fill=white] {\footnotesize Func. \& Graph.};
    \draw[|->] (X3) -- (Xp2) node[midway, centered, fill=white] {\footnotesize Func. \& Graph.};
\end{tikzpicture}
    \end{minipage}
    \vline
    \hspace{1mm}
    \begin{minipage}{0.52\textwidth}
    \centering
    \begin{tikzpicture}[scale=.8]
        \node at (0,1) {\large$\model$:};
        \node at (6.7,1) {\large$\model'$:};
        \node (p1) at (-.2, -0.3) {\footnotesize $P(Y)$};
        \node (p2) at (0.2, -1.7) {\footnotesize$P(Y|X)$};
        \node (p3) at (-0.2, -3.5) {\footnotesize$P(Y|X)$};
        \node (p1-2) at (6.4, -0.3) {\footnotesize$P(Y')$};
        \node (p2-2) at (7.15, -1.6) {\footnotesize$P(Y'|X')$};
        \node (p3-alpha) at (7, -2.7) {\footnotesize$P(Y'|\alpha_{X}(X))$};
        \node (alpha-p3) at (6.7, -4) {\footnotesize$\boldsymbol{\alpha} \left[P(Y|X)\right]$};
    
        \draw[dotted] (0,-2.25) ellipse (1.3cm and 2.5cm);
        \draw[dotted] (6.8,-2.25) ellipse (1.8cm and 2.5cm);
    
        \path
          (p1) edge[] node[fill=white](eq1) {\footnotesize=} (p2)
          (p1-2) edge[] node[fill=white](eq2) {\footnotesize=} (p2-2)
          (alpha-p3) edge[] node[fill=white](eq3) {\footnotesize=} (p3-alpha)
          (eq1) edge[<-|] node[below] {\footnotesize implies} node[above] {\footnotesize Graphical consistency} (eq2)
          (p3) edge[|->] node[below] {\footnotesize implies} node[above] {\footnotesize Functional consistency} (eq3);
    \end{tikzpicture}
    \end{minipage}
    \caption{Visualization of abstractions \textbf{(a)} and consistency \textbf{(b)}. Figure \textbf{(a)} illustrates the overlap and the differences between graphical and functional abstractions: both involve mapping and clustering variables; however, graphical abstractions preserve edges, while functional abstractions exploit the more detailed models to map variable ranges.
    Figure \textbf{(b)}: given an SCM $\model$ and an abstraction $\model'$, graphical consistency implies preservation of equality relations between distributions under abstraction, whereas functional consistency guarantees that abstraction and $\li$ operations (in this case conditioning) commute, yielding the same resulting distributions.}
    \label{fig:graphical-functional-consistency}
\end{figure}

However, graphical and functional consistency are intrinsically different, as one is defined on a graphical level and the other on a functional level. Indeed, they both require different identities in order to hold, as shown in Fig.\ref{fig:graphical-functional-consistency}:
\emph{graphical consistency} requires that an equality in the CDAG underlying an abstracted model $\model'$ hold in the original model $\model$;
\emph{functional consistency} implies the identity of two interventional distributions in the abstracted model $\model'$. We show that it is possible to align functional consistency with graphical consistency as follows:
\begin{proposition}\label{prop:graphical-implies-functional-consistency}
    A functional $\ltwo$-consistent abstraction $\boldsymbol{\alpha}: \model \rightarrow \model'$ with bijective maps $\alpha_{V}$ implies graphical $\ltwo$-consistency.
\end{proposition}

This proposition relies on the following lemma which guarantees that a functional abstraction $\boldsymbol{\alpha}: \model \rightarrow \model'$, with all range mappings $\alpha_{V}$ bijective, preserves equalities and inequalities. 

\begin{lemma}[Bijective Range Maps Preserve Distribution (In)Equalities] \label{lem:consistency-Equalities}
    \sloppy Let $\boldsymbol{\alpha}: \model \rightarrow \model'$ be an $\ltwo$-consistent abstraction with $\alpha_V:\domain\left(\varphi^{-1}(V)\right)\rightarrow\domain\left(V\right)$ bijective for all $V\in\vars_{\model'}$. Let $\mathbf X, \mathbf Y_1,\mathbf Y_2, \mathbf Z_1,\mathbf Z_2 \subseteq \vars_{\model}$ be partitions defined by $\varphi$, then $P(\mathbf X\:|\:do(\mathbf Y_1), \mathbf Z_1) = P(\mathbf X\:|\:do(\mathbf Y_2), \mathbf Z_2)$ if and only if $\alpha_{\mathbf{X}}[P(\mathbf X\:|\:do(\mathbf Y_1), \mathbf Z_1)] = \alpha_{\mathbf{X}}[P(\mathbf X\:|\:do(\mathbf Y_2), \mathbf Z_2)]$. \emph{[Proof in App. \ref{proof-consistency-equalities}.]}
\end{lemma}

The proof of Lem.\ref{lem:consistency-Equalities} shows that surjectivity of $\alpha_{V}$ is sufficient to imply that inequalities in $\model'$ must have a corresponding inequality in $\model$, while bijectivity ensures also that all equalities in $\model'$ have a corresponding equality in $\model$. 

\section{Aligning Graphical and Functional Abstraction}

After relating the notions of consistency, we now consider how graphical clustering and functional abstraction themselves are related. To unify the notation, we will use $V_i\in \mathbf V$ for variables in the base model $\model$ and $\mathbf C_i \in \mathbf{C}$ for variables in the abstracted model $\model'$, both in case of a graphical or functional abstraction. Now, as by Def.\ref{def:CDAG}, CDAGs have two important limitations: (i) they require every variable in $\model$ to belong to a cluster; and (ii) they specify an aggregation over variables in $\model$, but not over their values. Correspondingly, an aligned $\alpha$-abstraction must: (i) consider all variables as relevant, $\mathbf R = \vars$; and, (ii) have \emph{bijective range mappings}, as in Lem.\ref{lem:consistency-Equalities}. Under these two conditions, we now show a correspondence between bijective $\ltwo$-consistent $\alpha$-abstractions with $\mathbf R = \mathbf{V}$ and the set of CDAGs. Note that for applications the restrictions of the CDAG may impede the use of it for real world abstractions. We will illustrate this using an example in the next section.

In the direction $\alpha$-abstraction $\Rightarrow$ CDAG, we prove that a bijective $\ltwo$-consistent $\alpha$-abstraction with $\mathbf R = \mathbf{V}$ entails a unique CDAG. 
To do so, we show that an $\ltwo$-consistent $\alpha$-abstraction defines a structure which complies with the two constructive rules of a CDAG in Def.\ref{def:CDAG}. Rule 1 is satisfied by showing that the mapping of variables $\varphi: \mathbf V \rightarrow \mathbf C$ of an $\ltwo$-consistent $\alpha$-abstraction uniquely determines the causal edges of $\model'$ as a CDAG would:

\begin{lemma}[$\ltwo$-Consistency Uniquely Determines Adjacencies] \label{lem:interventional-consistency-adjacency}
    Given an $\ltwo$-consistent bijective abstraction $\boldsymbol{\alpha}: \mathcal M\rightarrow \mathcal M'$, adjacencies between variables in $\mathcal M'$ are uniquely determined by the map $\varphi: \mathbf V \rightarrow \mathbf C$ and comply with Rule 1 of Def.\ref{def:CDAG} (CDAG). \emph{[Proof in App. \ref{proof-preservation-adjacencies}.]}
\end{lemma}

The proof of Lem.\ref{lem:interventional-consistency-adjacency} shows that surjectivity alone guarantees that a causal edge in the abstraction implies a compatible causal edge in the base model. 
Conversely, for any bijective $\ltwo$-consistent $\alpha$-abstraction, if there exists an adjacency in the base model, there must exists a corresponding adjacency in the abstraction.

Next we prove that Rule 2 is also satisfied by showing that confounding edges are similarly uniquely defined by an $\ltwo$-consistent $\alpha$-abstraction in accordance with the CDAG definition:

\begin{lemma}[$\ltwo$-Consistency Uniquely Determines Confounding Edges]\label{lem:L2-conf}
    Given an $\ltwo$-consistent bijective abstraction $\boldsymbol{\alpha}: \mathcal M\rightarrow \mathcal M'$, confounding edges between variables in $\mathcal M'$ are uniquely determined by the map $\varphi: \mathbf V \rightarrow \mathbf C$ and comply with Rule 2 of Def.\ref{def:CDAG} (CDAG). \emph{[Proof in App. \ref{proof-L2-confounding}.]}
\end{lemma}

Determining adjacencies according to Rule 1 and confounding according to Rule 2 immediately implies a unique CDAG structure:

\begin{theorem}[$\alpha$-Abstraction $\Rightarrow$ CDAG]\label{thm:CDAG-alpha}
     \sloppy Given any bijective $\ltwo$-consistent $\alpha$-abstraction $\boldsymbol{\alpha}: \model\rightarrow\model'$ with $\relevant = \vars$,  the DAG of $\model'$ is a CDAG of the graph of $\model$. \emph{[Proof in App. \ref{proof-cdag-alpha}.]}
\end{theorem}

This theorem uncovers a clear connection between graphical and functional abstraction. Despite the fact that the definition of $\alpha$-abstraction makes \emph{no} reference at all to the graphical structure of an abstracted model, this theorem shows that for a bijective $\alpha$-abstraction, if we want to guarantee $\ltwo$-consistency, there is a \emph{unique} admissible graphical structure; and this structure is exactly the one derived using the CDAG construction rules.

In the direction CDAG $\Rightarrow$ $\alpha$-abstraction, we can show that for every CDAG there is an immediate bijective $\ltwo$-consistent $\alpha$-abstraction such that the underlying structure is that CDAG.
\begin{theorem}[CDAG $\Rightarrow$ $\alpha$-Abstraction]\label{thm:alpha-cdag}
     \sloppy Given a CDAG there exists an equivalent $\ltwo$-consistent $\alpha$-abstraction $\boldsymbol{\alpha}:\model\rightarrow\model'$ with $\relevant = \vars$, $\varphi$ given by the clustering, and all maps $\alpha_{\clusts}$ bijective. \emph{[Proof in App. \ref{proof-alpha-cdag}.]}
\end{theorem}

Thm.\ref{thm:CDAG-alpha} and Thm.\ref{thm:alpha-cdag} establish a correspondence  between the set of bijective $\ltwo$-consistent $\alpha$-abstraction and the set of CDAGs. However, notice there might still exist non-bijective $\ltwo$-consistent $\alpha$-abstraction or even non-$\ltwo$-consistent $\alpha$-abstraction with an underlying CDAG. For intuition about the bijectivity assumption, consider Example \ref{app:example-sur} in the appendix.

\section{Extending Graphical Abstractions: Partial CDAGs}

Graphical abstractions in the form of CDAGs are bound to account in clustering for all variables ($\relevant = \vars$). This restriction limits the expressivity and usability of graphical clustering as illustrated in the following example.

\begin{example}\label{ex:motivating}
    \normalfont Let $\mathcal{M}: \langle\mathbf{U}, \mathbf{V}, \mathcal{F}, P(\mathbf U)\rangle$ be an SCM modeling the effects of smoking ($X_1$) and air pollution ($X_2$) on lung cancer ($Y_1$) and shortness of breath ($Y_2$), through the mediating variable of tar deposits ($Z$). We assume $X_1$ and $X_2$ to be independent, and $Y_1$ and $Y_2$ to be independent given $Z$, as shown in the DAG of Fig.\ref{fig:ex-1} \textbf{(a)}.   
    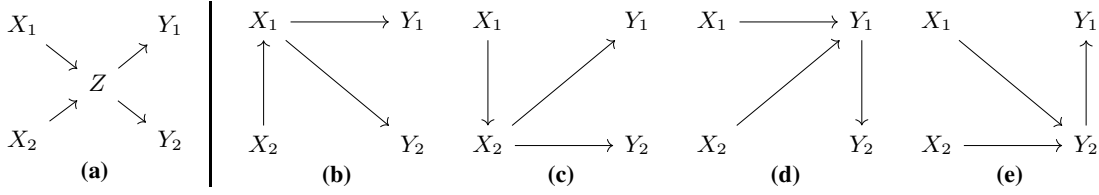
\begin{figure}[t]
        \centering
        \begin{minipage}{0.2\textwidth}\footnotesize
            \centering
            \vspace*{\fill}
            \begin{tikzcd}[row sep=small, column sep=small]
                X_1 \arrow[dr] & & Y_1 \\       
                & Z \arrow[ur] \arrow[dr] &\\   
                X_2 \arrow[ur] & & Y_2          
            \end{tikzcd}\\
            \textbf{ (a)}
            \vspace*{\fill}
        \end{minipage}%
        \vline
        \hspace{0.0\textwidth}
        \begin{minipage}{0.79\textwidth}\footnotesize
            \centering
            \begin{minipage}{0.24\textwidth}
                \centering
                \begin{tikzcd}[row sep=.45cm, column sep=small]
                    X_1 \arrow[ddrr]\arrow[rr] & & Y_1\\
                    &\;\;\; & \\
                    X_2 \arrow[uu] & & Y_2
                \end{tikzcd}\\
                \textbf{(b)}
            \end{minipage}\hspace{0.00\textwidth}
            \begin{minipage}{0.24\textwidth}
                \centering
                \begin{tikzcd}[row sep=.45cm, column sep=small]
                    X_1 \arrow[dd] & & Y_1\\
                    &\;\;\; & \\
                    X_2 \arrow[uurr]\arrow[rr] & & Y_2
                \end{tikzcd}\\
                \textbf{(c)}
            \end{minipage}
            \begin{minipage}{0.24\textwidth}
                \centering
                \begin{tikzcd}[row sep=.45cm, column sep=small]
                    X_1 \arrow[rr] & & Y_1 \arrow[dd]\\
                    &\;\;\; & \\
                    X_2 \arrow[uurr] & & Y_2
                \end{tikzcd}\\
                \textbf{(d)}
            \end{minipage}\hspace{0.0\textwidth}
            \begin{minipage}{0.24\textwidth}
                \centering
                \begin{tikzcd}[row sep=.45cm, column sep=small]
                    X_1 \arrow[ddrr] & & Y_1\\
                    &\;\;\; & \\
                    X_2 \arrow[rr] & & Y_2 \arrow[uu]
                \end{tikzcd}\\
                \textbf{(e)}
            \end{minipage}
        \end{minipage}
        
        \caption{Given the DAG representing an SCM \textbf{(a)}, there are 4 ways of clustering such that $X_1$, $X_2$, $Y_1$, and $Y_2$ are in separate clusters : $\{\{X_1, Z\},\{X_2\},\{Y_1\},\{Y_2\}\}$ \textbf{(b)}, $\{\{X_1\},\{X_2, Z\},\{Y_1\},\{Y_2\}\}$ \textbf{(c)}, $\{\{X_1\},\{X_2\},\{Y_1, Z\},\{Y_2\}\}$ \textbf{(d)}, $\{\{X_1\},\{X_2\},\{Y_1\},\{Y_2, Z\}\}$ \textbf{(e)}. Observe that CDAGs \textbf{(b)} and \textbf{(c)} lose the ability to intervene on $X_1$ and $X_2$ independently, while \textbf{(d)} and \textbf{(e)} retain the ability to intervene independently on $X_1$ and $X_2$, but make $Y_1$ a direct cause of $Y_2$ or vice versa.}
        \label{fig:ex-1}
    \end{figure}
    Assume we are interested in the effects of $X_1,X_2$ (smoking and air quality) on $Y_1,Y_2$ (lung cancer and shortness of breath), and we have no way of measuring the tar deposits $Z$. 
    If we want to keep the possibility of intervening on $X_1$ or $X_2$ independently, and the ability of predicting $Y_1$ or $Y_2$ independently, any CDAG abstracting away $Z$ as in Fig.\ref{fig:ex-1} will sacrifice one of these possibilities: either our ability to intervene (subfigures \textbf{(b)}, \textbf{(c)}) or to independently predict (subfigures \textbf{(d)}, \textbf{(e)}).
    The structure of the CDAGs does not seem to align to the description of two causes and two effects because of the need to include $Z$ in one of the clusters. Ideally, we want to create an abstraction removing $Z$ that preserves our ability to intervene on $X_1, X_2$, predictability of $Y_1,Y_2$, and any confounding effect introduced by $Z$.
\end{example}

To overcome the limitation described in the example, we now extend the expressivity of graphical models by defining Partial CDAGs:

\begin{definition} [Partial CDAG]\label{def:PCDAG} 
    Let $G = \langle \mathbf{V},\mathbf E\rangle$ be a DAG and $\varphi: \vars \rightarrow \mathbf{C}$ be a partial surjective function where $\mathbf C = \{\mathbf C_1, \dots, \mathbf C_k\}$. Let $\remain \subseteq \vars$ be the remainder set of vertices that are not mapped by $\varphi$. $\mathbf{C} \cup \remain$ forms a partition of the vertices $\vars$. $G_{\mathbf C} = \langle \mathbf C, \mathbf E_{\mathbf C}\rangle $ is a partial CDAG (PCDAG) of $G$ if and only if the set of edges $\mathbf E_{\mathbf C}$ abides by the following rules:
    
    \begin{enumerate}
        \item An edge $\mathbf C_i \rightarrow \mathbf C_j$ is in $\mathbf E_{\mathbf C}$ if there exists some path $V_i\rightarrow \dots \rightarrow V_j$ with zero or more intermediate variables in $\mathbf V$, such that $ V_i\in \mathbf C_i,\; V_j\in \mathbf C_j,\:\; \clusts_i \neq \clusts_j$, and all intermediate variables  are in $\remain$.
        \item[2A.] A bidirected (confounding) edge $\mathbf C_i \confarrow \mathbf C_j$ is in $\mathbf E_{\mathbf C}$ if there exists some {$V_i \nolinebreak \leftarrow \nolinebreak \dots\nolinebreak  \confarrow\nolinebreak  \dots \nolinebreak \rightarrow \nolinebreak V_j$} with zero or more intermediate variables on either side in $\mathbf V$, such that $V_i \in \mathbf C_i, \; V_j \in \mathbf C_j ,\:\; \clusts_i \neq \clusts_j,$ and all intermediate variables are in $\remain$.
        \item[2B.] A bidirected (confounding) edge $\mathbf C_i \confarrow \mathbf C_j$ is in $\mathbf E_{\mathbf C}$ if there exists some $Q \in \remain$ with paths $V_i \leftarrow \dots \leftarrow Q \rightarrow \dots \rightarrow V_j$ and $V_i\in \mathbf C_i,\; V_j \in \mathbf C_j, \; \clusts_i \neq \clusts_j$, such that all intermediate vertices in the paths $Q \rightarrow\dots \rightarrow V_i$ and $Q \rightarrow\dots \rightarrow V_j$ are in $\remain$.
    \end{enumerate}
    Further, it is required that the graph $G_{\clusts}$ is induced by $\varphi$ acyclic.
\end{definition}

The first two rules follow intuitively as extensions of CDAGs: Rule 1 ensures that directed paths and adjacencies are maintained, while Rule 2A ensures preservation of confounding edges in PCDAGs. However, by allowing for partial clustering more confounding edges might be introduced; therefore, Rule 2B is required to capture confounding edges introduced when shared parents are dropped. To understand the role of 2B see the following example.

\begin{example} \normalfont 
    We continue with the SCM of Example \ref{ex:motivating}, but we now define a partial clustering $\clusts = \{\{X_1\}, \{X_2\}, \{Y_1\}, \{Y_2\}\}$ with remainder set $\remain = \{Z\}$. By applying only Rule 1 and Rule 2A we would obtain the second model in Fig.\ref{fig:ex-2}. However, notice that in the original model it holds that $\nolinebreak{Y_1 \nind  Y_2\;|\;X_1, X_2}$ due to $Z$. In the proposed PCDAG this dependence does not hold. In order to guarantee that $\nolinebreak{\{Y_1\} \nind  \{Y_2\}\;|\;\{X_1\}, \{X_2\}}$, a confounding arrow $\{Y_1\}\confarrow\{Y_2\}$ must be introduced as by Rule 2B. The last DAG of Fig.\ref{fig:ex-2} shows the PCDAG abstraction.
    
    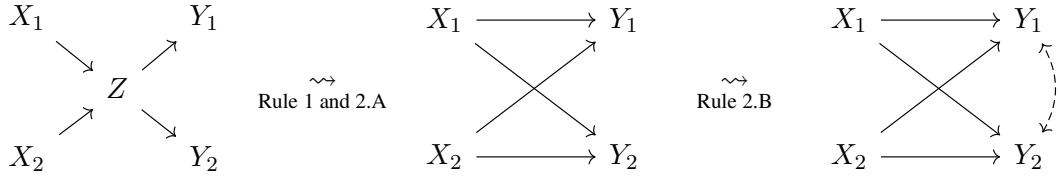
\begin{figure}[]
        \centering
        \begin{tikzcd}[row sep=small, column sep=small]
            X_1 \arrow[dr] & & Y_1 \\       
            & Z \arrow[ur] \arrow[dr] &\\   
            X_2 \arrow[ur] & & Y_2          
        \end{tikzcd}
        \hspace{.25cm}$\underset{\text{Rule 1 and 2.A}}{\rightsquigarrow}$\hspace{.25cm}
        \begin{tikzcd}[row sep=.45cm, column sep=small]
            X_1 \arrow[ddrr]\arrow[rr] & & Y_1\\
            &\;\;\; & \\
            X_2 \arrow[uurr]\arrow[rr] & & Y_2
        \end{tikzcd}
        \hspace{.5cm}$\underset{\text{Rule 2.B}}{\rightsquigarrow}$\hspace{.5cm}
        \begin{tikzcd}[row sep=.45cm, column sep=small]
            X_1 \arrow[ddrr]\arrow[rr] & & Y_1\\
            &\;\;\; & \\
            X_2 \arrow[uurr]\arrow[rr] & & Y_2 \arrow[uu, bend right, dashed, leftrightarrow]  
        \end{tikzcd}
        \caption{Given an SCM represented by the left DAG, there exists an abstraction that preserves marginal independence $X_1 \ind X_2$, and independent predictability of $Y_1$ and $Y_2$ given by the PCDAG. Notice how the application of rules 1 and 2A preserve causal and confounding edges from the original model, whereas 2B introduces confounding edges preserving confounding effects of removed fork structures.}
        \label{fig:ex-2}
    \end{figure} 
\end{example}

PCDAGs allow for dropping variables while keeping their confounding effects; this offers more versatility as, in the case of the example above, it allows us to preserve independent predictability (over $Y_1,Y_2$) \textit{and} independent interventions (over $X_1,X_2$).

Similarly to CDAGs, we now prove some key graphical properties of PCDAGs related to the preservation of adjancencies and directed paths. First of all, in PCDAGs we require a more flexible notion of adjacency to account for dropped variables:

\begin{definition}[Mediated Adjacency]\label{def:med-adj}
    Let $\mathcal M$ be an SCM with a PCDAG defined by clusters $\mathbf C$ and remainder set $\remain$. Given two variables $V_i,V_j\in\vars$, there exists a mediated adjacency $V_i\rightsquigarrow V_j$ if and only if there exists a directed path $V_i\rightarrow\dots\rightarrow V_j$ such that all intermediate variables are in $\remain$.
\end{definition}

This concept of mediated adjacency is equivalent to \textbf{T}-direct path in linear abstraction \citep{massidda2024learningcausalabstractionslinear}. PCDAGs preserve mediated adjacencies by construction:
\begin{lemma}[PCDAGs Preserve Mediated Adjancencies]\label{lem:PCDAG-Adjancency}
    Let $\mathcal M$ be an SCM with a PCDAG defined by clusters $\mathbf C$ and remainder set $\remain$. Let $\mathbf C_i, \mathbf C_j \in \mathbf C, \; \mathbf C_i \neq \mathbf C_j$ and $V_i\in \mathbf C_i,\;V_j\in \mathbf C_j$, then a mediated adjacency between $V_i$ and $V_j$ exists if and only if there exists an adjacency between $\mathbf C_i$ and $\mathbf C_j$. \emph{[Proof in App. \ref{proof-PCDAG-adj}.]}
\end{lemma}

Preservation of mediated adjacencies immediately implies preservation of directed paths, as any directed path can be decomposed into a series of consecutive adjacencies:

\begin{lemma}[PCDAGs Preserve Directed Paths]\label{lem:PCDAG-Directed-Paths}
    Let $\mathcal M$ be an SCM with a PCDAG defined by clusters $\mathbf C$ and remainder set $\remain$. Let $\mathbf C_i, \mathbf C_j \in \mathbf C, \; \mathbf C_i \neq \mathbf C_j$ and $V_i\in \mathbf C_i,\;V_j\in \mathbf C_j$, then a directed path $\mathbf C_i \rightarrow \dots \rightarrow \mathbf C_j$ exists if there exists a directed path $V_i \rightarrow \dots \rightarrow V_j$. \emph{[Proof in App. \ref{proof-PCDAG-dir-paths}.]}
\end{lemma}

We thus align ourselves with the results of \cite{anand2023cdag} by showing that PCDAGs preserve adjacencies and directed paths. This in turn allows us to discuss the causal consistency of PCDAGs and show that a PCDAG is compatible with functional $\ltwo$-consistency:

\begin{theorem}[$\ltwo$-consistency of PCDAGs]\label{thm:PCDAG-L2-Consistent}
    Given an SCM $\model$ and a PCDAG $\model'$ of $\model$, there exists a set of mechanisms $\mathcal F_{\model'}$ such that $\model'$ is $\ltwo$-consistent with $\model$. \emph{[Proof in App. \ref{proof-PCDAG-Consistency}.]}
\end{theorem}

As PCDAGs can be compatible with functional $\ltwo$-consistency, Prop.\ref{prop:graphical-implies-functional-consistency} states that PCDAGs are necessarily graphically $\ltwo$-consistent.

\section{Re-Aligning Graphical and Functional Abstraction}
    
We now extend our previous results to the alignment of graphical abstraction in the form of PCDAGs and functional abstractions.  
Differently from a CDAG, a PCDAG has only one limitation compared to $\alpha$-abstractions: (i) it specifies an aggregation over variables in $\model$, but not over the values of the variables. 
Therefore, a corresponding $\alpha$-abstraction only needs to (i) have \emph{bijective range mappings}. With this condition, we now show a correspondance between bijective $\ltwo$-consistent $\alpha$-abstraction (without necessarily $\mathbf{R}=\vars$) and the set of PCDAGs.

In the direction $\alpha$-abstraction $\Rightarrow$ PCDAG, we prove that a bijective $\ltwo$-consistent $\alpha$-abstraction entails a unique PCDAG. 
As before, we show that a bijective $\ltwo$-consistent $\alpha$-abstraction defines a structure that complies with the three constructive rules of a PCDAG in Def.\ref{def:PCDAG}. 
Rule 1 is satisfied by showing that the mapping of variables $\varphi: \mathbf{V} \rightarrow \mathbf{C}$ of an $\ltwo$-consistent $\alpha$-abstraction uniquely determines all the causal edges of $\model'$ as a PCDAG would:

\begin{lemma}[$\ltwo$-Consistency Uniquely Determines Adjacencies]
    \label{lem:interventional-consistency-adjacency-2}
    Given a bijective $\ltwo$-consistent abstraction $\boldsymbol{\alpha}: \mathcal M\rightarrow \mathcal M'$, adjacencies between variables in $\mathcal M'$ are uniquely determined by the map $\varphi: \mathbf V \rightarrow \mathbf C$ and comply with Rule 1 of Def.\ref{def:PCDAG} (PCDAG). \emph{[Proof in App. \ref{proof-preservation-adjacencies-2}.]} \looseness=-1
\end{lemma}

Next we prove that confounding edges of a bijective $\ltwo$-consistent abstraction are necessarily equal to those given by Rule 2A and 2B of a PCDAG in Def.\ref{def:PCDAG}:

\begin{lemma}[$\ltwo$-Consistency Uniquely Determines Confounding Edges]\label{lem:L2-conf-2}
    Given a bijective $\ltwo$-consistent abstraction $\boldsymbol{\alpha}: \mathcal M\rightarrow \mathcal M'$, confounding edges between variables in $\mathcal M'$ are uniquely determined by the map $\varphi: \mathbf V \rightarrow \mathbf C$ and comply with Rule 2A and 2B of Def.\ref{def:PCDAG} (PCDAG). \emph{[Proof in App. \ref{proof-L2-confounding-2}.]}
\end{lemma}

Determining mediated adjacencies according to Rule 1 and confounding according to Rule 2A and 2B immediately implies a unique PCDAG structure:

\begin{theorem}[$\alpha$-Abstraction $\Rightarrow$ PCDAG]\label{thm:alpha-PCDAG}
     \sloppy Given any bijective $\ltwo$-consistent $\alpha$-abstraction $\boldsymbol{\alpha}: \model\rightarrow\model'$, the DAG of $\model'$ is a PCDAG of the graph of $\model$. \emph{[Proof in App. \ref{proof-pcdag-alpha}.]}
\end{theorem}

And, mirroring the results of the CDAG, in the direction PCDAG $\Rightarrow$ $\ltwo$-consistent $\alpha$-abstraction we get that for all PCDAGs we can construct an equivalent $\ltwo$-consistent $\alpha$-abstraction by taking the range maps $\alpha_\clusts$ to be bijective.
\begin{theorem}[PCDAG $\Rightarrow$ $\alpha$-Abstraction]\label{thm:PCDAG-alpha}
     \sloppy Given a PCDAG there exists an equivalent $\ltwo$-consistent $\alpha$-abstraction $\boldsymbol{\alpha}:\model\rightarrow\model'$ with $\relevant \subseteq \vars$, $\varphi$ given by the clustering, and all maps $\alpha_{\clusts}$ bijective. \emph{[Proof in App. \ref{proof-alpha-PCDAG}.]}
\end{theorem}

Furthermore, by Prop.\ref{prop:graphical-implies-functional-consistency}, the PCDAG is necessarily graphically $\ltwo$-consistent. Thus, if we want a bijective $\ltwo$-consistent abstraction, we can start by applying the rules of the PCDAG. Moreover, by the proof of Lem.\ref{lem:interventional-consistency-adjacency} we get that the PCDAG can also be useful for surjective $\ltwo$-consistent abstractions, albeit that not all edges in the PCDAG may be necessary. In other words, PCDAGs can also describe the structure of surjective $\ltwo$-consistent abstractions, but may lose faithfulness.

\section{Relation to other functional abstractions}

Another alternative functional abstraction approach is given by $\tau$-$\omega$ framework \citep{rubenstein2017causal}; see App. \ref{app:def-tau-omega} for a definition of a $\tau$-$\omega$ abstraction. Here we show that a particular well-behaved form of $\tau$-$\omega$ abstraction called constructive $\tau$-abstraction (\cite{beckers2019abstracting}) is equivalent to an $\alpha$-abstraction. This equivalence immediately allows us to extend the connection between PCDAGs and bijective constructive $\tau$-abstractions. 

\begin{corollary}[Equivalence $\alpha$-abstraction and Constructive $\tau$-abstraction]\label{cor:alpha-constructive-tau}
    \sloppy The $\alpha$-abstraction is equivalent to the constructive $\tau$-abstraction, if for all settings $\mathbf{v}\in\domain(\vars)$ there exists a $\mathbf{u}\in\domain(\exos)$ giving rise to $\mathbf{v}$. \emph{[Proof in App. \ref{proof-alpha-constructive-tau}.]}
\end{corollary}

\section{Conclusion}
    In this paper we have shown how graphical and functional abstraction are related by showing the connection between CDAGs/PCDAGs and bijective $\ltwo$-consistent $\alpha$-abstractions. This alignment highlights the dual graphical and functional nature of abstractions (similar to SCMs) and allows us to take advantage both of the constructive definitions of graphical abstractions (which can provide strong consistency guarantees by construction) and the declarative definitions of functional abstractions (which establish explicit maps between the models and the data generated by the models). Furthermore, alignment of graphical and functional abstraction suggests that any functional abstraction learning algorithm aimed at learning new simplified models with the requirement of $\ltwo$-consistency can rely on the algorithmic procedure in the definition of the PCDAG to learn the structure of an abstracted model. Future work will consider further theoretical study of the relations between the abstraction frameworks, in particular considering implicit and explicit restrictions and their implications; this would allow for a more immediate transfer of results and methods across frameworks. Finally, our results may be exploited to improve abstraction learning algorithm by taking advantage of graphical and functional aspects.

\bibliography{refs}

\begin{thebibliography}{20}
\providecommand{\natexlab}[1]{#1}
\providecommand{\url}[1]{\texttt{#1}}
\expandafter\ifx\csname urlstyle\endcsname\relax
  \providecommand{\doi}[1]{doi: #1}\else
  \providecommand{\doi}{doi: \begingroup \urlstyle{rm}\Url}\fi

\bibitem[Anand et~al.(2023)Anand, Ribeiro, Tian, and Bareinboim]{anand2023cdag}
Tara~V Anand, Adele~H Ribeiro, Jin Tian, and Elias Bareinboim.
\newblock Causal effect identification in cluster dags.
\newblock In \emph{Proceedings of the AAAI Conference on Artificial Intelligence}, volume~37, pages 12172--12179, 2023.

\bibitem[Bareinboim et~al.(2022)Bareinboim, Correa, Ibeling, and Icard]{Bareinboim2020hierarchy}
Elias Bareinboim, Juan~D. Correa, Duligur Ibeling, and Thomas Icard.
\newblock \emph{On Pearl’s Hierarchy and the Foundations of Causal Inference}, page 507–556.
\newblock Association for Computing Machinery, New York, NY, USA, 1 edition, 2022.
\newblock ISBN 9781450395861.
\newblock URL \url{https://doi.org/10.1145/3501714.3501743}.

\bibitem[Beckers and Halpern(2019)]{beckers2019abstracting}
Sander Beckers and Joseph~Y Halpern.
\newblock Abstracting causal models.
\newblock In \emph{Proceedings of the aaai conference on artificial intelligence}, volume~33, pages 2678--2685, 2019.

\bibitem[Dyer et~al.(2023)Dyer, Bishop, Felekis, Zennaro, Calinescu, Damoulas, and Wooldridge]{dyer2023interventionally}
Joel Dyer, Nicholas Bishop, Yorgos Felekis, Fabio~Massimo Zennaro, Anisoara Calinescu, Theodoros Damoulas, and Michael Wooldridge.
\newblock Interventionally consistent surrogates for agent-based simulators.
\newblock \emph{arXiv preprint arXiv:2312.11158}, 2023.

\bibitem[Felekis et~al.(2024)Felekis, Zennaro, Branchini, and Damoulas]{felekis2023causal}
Yorgos Felekis, Fabio~Massimo Zennaro, Nicola Branchini, and Theodoros Damoulas.
\newblock Causal optimal transport of abstractions.
\newblock In \emph{Causal Learning and Reasoning}, pages 462--498. PMLR, 2024.

\bibitem[Geiger et~al.(2021)Geiger, Lu, Icard, and Potts]{geiger2021causal}
Atticus Geiger, Hanson Lu, Thomas Icard, and Christopher Potts.
\newblock Causal abstractions of neural networks.
\newblock \emph{Advances in Neural Information Processing Systems}, 34:\penalty0 9574--9586, 2021.

\bibitem[Geiger et~al.(1990)Geiger, Verma, and Pearl]{geiger1990d}
Dan Geiger, Thomas Verma, and Judea Pearl.
\newblock d-separation: From theorems to algorithms.
\newblock In \emph{Machine intelligence and pattern recognition}, volume~10, pages 139--148. Elsevier, 1990.

\bibitem[Keki{\'c} et~al.(2024)Keki{\'c}, Sch{\"o}lkopf, and Besserve]{kekic2023targeted}
Armin Keki{\'c}, Bernhard Sch{\"o}lkopf, and Michel Besserve.
\newblock Targeted reduction of causal models.
\newblock In \emph{Uncertainty in Artificial Intelligence}, pages 1953--1980. PMLR, 2024.

\bibitem[Massidda et~al.(2022)Massidda, Geiger, Icard, and Bacciu]{massidda2022causal}
Riccardo Massidda, Atticus Geiger, Thomas Icard, and Davide Bacciu.
\newblock Causal abstraction with soft interventions.
\newblock \emph{arXiv preprint arXiv:2211.12270}, 2022.

\bibitem[Massidda et~al.(2024)Massidda, Magliacane, and Bacciu]{massidda2024learningcausalabstractionslinear}
Riccardo Massidda, Sara Magliacane, and Davide Bacciu.
\newblock Learning causal abstractions of linear structural causal models.
\newblock In \emph{Uncertainty in Artificial Intelligence}, pages 2486--2515. PMLR, 2024.

\bibitem[Parviainen and Kaski(2017)]{PARVIAINEN2017110}
Pekka Parviainen and Samuel Kaski.
\newblock Learning structures of bayesian networks for variable groups.
\newblock \emph{International Journal of Approximate Reasoning}, 88:\penalty0 110--127, 2017.
\newblock ISSN 0888-613X.
\newblock \doi{https://doi.org/10.1016/j.ijar.2017.05.006}.
\newblock URL \url{https://www.sciencedirect.com/science/article/pii/S0888613X17303134}.

\bibitem[Pearl(2009)]{pearl2000models}
Judea Pearl.
\newblock \emph{Causality: Models, Reasoning and Inference}.
\newblock Cambridge University Press, USA, 2nd edition, 2009.
\newblock ISBN 052189560X.

\bibitem[Rischel and Weichwald(2021)]{rischel2021compabstraction}
Eigil~F. Rischel and Sebastian Weichwald.
\newblock Compositional abstraction error and a category of causal models.
\newblock In Cassio de~Campos and Marloes~H. Maathuis, editors, \emph{Proceedings of the Thirty-Seventh Conference on Uncertainty in Artificial Intelligence}, volume 161 of \emph{Proceedings of Machine Learning Research}, pages 1013--1023. PMLR, 27--30 Jul 2021.
\newblock URL \url{https://proceedings.mlr.press/v161/rischel21a.html}.

\bibitem[Rischel(2020)]{rischel2020category}
Eigil~Fjeldgren Rischel.
\newblock The category theory of causal models.
\newblock \emph{Master's thesis, University of Copenhagen}, 2020.

\bibitem[Rubenstein et~al.(2017)Rubenstein, Weichwald, Bongers, Mooij, Janzing, Grosse-Wentrup, and Sch{\"o}lkopf]{rubenstein2017causal}
Paul~K Rubenstein, Sebastian Weichwald, Stephan Bongers, Joris~M Mooij, Dominik Janzing, Moritz Grosse-Wentrup, and Bernhard Sch{\"o}lkopf.
\newblock Causal consistency of structural equation models.
\newblock \emph{33rd Conference on Uncertainty in Artificial Intelligence 2017}, 2017.

\bibitem[Schölkopf et~al.(2021)Schölkopf, Locatello, Bauer, Ke, Kalchbrenner, Goyal, and Bengio]{scholkopf2021crlsurvey}
Bernhard Schölkopf, Francesco Locatello, Stefan Bauer, Nan~Rosemary Ke, Nal Kalchbrenner, Anirudh Goyal, and Yoshua Bengio.
\newblock Toward causal representation learning.
\newblock \emph{Proceedings of the IEEE}, 109\penalty0 (5):\penalty0 612--634, 2021.
\newblock \doi{10.1109/JPROC.2021.3058954}.

\bibitem[Wahl et~al.(2024)Wahl, Ninad, and Runge]{wahl2024foundationsgroup}
Jonas Wahl, Urmi Ninad, and Jakob Runge.
\newblock Foundations of causal discovery on groups of variables.
\newblock \emph{Journal of Causal Inference}, 12, 07 2024.
\newblock \doi{10.1515/jci-2023-0041}.

\bibitem[Xia and Bareinboim(2024)]{xia2024neuralabs}
Kevin Xia and Elias Bareinboim.
\newblock Neural causal abstractions.
\newblock In \emph{Proceedings of the AAAI Conference on Artificial Intelligence}, volume~38, pages 20585--20595, 2024.

\bibitem[Zennaro et~al.(2023)Zennaro, Dr{\'a}vucz, Apachitei, Widanage, and Damoulas]{zennaro2023jointly}
Fabio~Massimo Zennaro, M{\'a}t{\'e} Dr{\'a}vucz, Geanina Apachitei, W.~Dhammika Widanage, and Theodoros Damoulas.
\newblock Jointly learning consistent causal abstractions over multiple interventional distributions.
\newblock In \emph{2nd Conference on Causal Learning and Reasoning}, 2023.
\newblock URL \url{https://openreview.net/forum?id=RNs7aMS6zDq}.

\bibitem[Zennaro et~al.(2024)Zennaro, Bishop, Dyer, Felekis, Calinescu, Wooldridge, and Damoulas]{zennaro2024causally}
Fabio~Massimo Zennaro, Nicholas~George Bishop, Joel Dyer, Yorgos Felekis, Ani Calinescu, Michael~J Wooldridge, and Theodoros Damoulas.
\newblock Causally abstracted multi-armed bandits.
\newblock In \emph{The 40th Conference on Uncertainty in Artificial Intelligence}, 2024.

\end{thebibliography}

\appendix

\appendix

\section{Supplements}
\subsection{Finite Graphical Model and $\alpha$-abstraction}\label{app:original-alpha}
    This supplement defines the $\alpha$-abstraction as introduced by \cite{rischel2020category, rischel2021compabstraction}. First we must introduce the Finite Graphical Model: a formalism for causal models, different from the SCM.

\begin{definition}[Finite Graphical Model] \textup{(\cite{rischel2020category})} 
A finite graphical model $\model$ contains the following:
    \begin{enumerate}
        \item A Directed Acyclic Graph (DAG) $G=\langle \mathbf V,\mathbf E\rangle$.
        \item For each vertex $V\in \mathbf V$ a finite set of values it can take, denoted by $\model[V]$.
        \item For each vertex $V\in \mathbf V$ a stochastic matrix $\model[\psi_V]$ giving the probability distributions of $V$ for all values its parents, $Pa(V)$, can take.
    \end{enumerate}
\end{definition}

The Finite Graphical Model is the basis for the definition of the $\alpha$-abstraction. The $\alpha$-abstraction defines the maps and properties that describe an abstraction from a low level Finite Graphical Model to a high level one.

\begin{definition}[$\alpha$-abstraction] \textup{(\cite{rischel2020category, rischel2021compabstraction})}
    An abstraction of finite graphical models $\alpha: \mathcal M \rightarrow \mathcal M'$ consists of the following:
    \begin{enumerate}
        \item A subset $\mathbf R\subseteq \mathbf V_\mathcal M$ of relevant variables.
        \item A surjective map $\varphi: \mathbf R \rightarrow \mathbf V_{\mathcal M'}$. Mapping all relevant variables to the variables of the abstraction.
        \item For each $V \in \mathbf{V}_{\mathcal{M}'}$, a surjective function $\domain(\varphi^{-1}(V)) \rightarrow \domain(V)$. Mapping the range of the pre-image $\varphi^{-1}(V)$ in $\mathcal M$ to the range of $V$ in $\mathcal M'$.
    \end{enumerate}
\end{definition}

\subsection{Example - Surjectivity vs. CDAG}\label{app:example-sur}
    \begin{example}[Surjectivity does not imply a CDAG]\normalfont
    \sloppy Here we will construct an example showing that only surjectivity is not a sufficient assumption to ensure that a $\ltwo$-consistent \alphabs{} agrees with the structure of a CDAG, motivating the bijectivity assumption used throughout the paper. Note however, that bijectivity itself is a sufficient but not minimal assumption, as will become clear in this example.

    First consider the SCM $\model$ with variables $\vars_{\model}=\{X,Z,Y\}$, $\domain(X) = \{1,2\}$, $\domain(Z) = \{1,3,5,7,9,\dots\}$, and $f_Y\in\funcs_{\model} = X\times Z$, giving rise to the DAG:

    
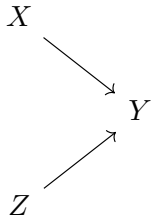
\begin{center}
        \begin{tikzcd}
            X \arrow[dr] & \\
            & Y\\
            Z\arrow[ur]& 
        \end{tikzcd}
    \end{center}

    Secondly, consider the SCM $\model'$ with variables $\vars_{\model'}=\{X',Z',Y'\}$, $\domain(X') = \{1,2\}$, $\domain(Z') = \{1,3,5,7,9,\dots\}$, and $f_{Y'}\in\funcs_{\model'} = X \times Z \mod 2$. It follows immediately that $Y'$ is dependent only on the value of $X$ as for all $z \in \domain(Z)$, $X\times z \mod 2$ equals 1 if X is 1, and 0 if $X$ is 2.

    There is a natural abstraction $\boldsymbol{\alpha}: \model\rightarrow\model'$ given by the maps:
    \begin{align*}
        \varphi&:=\{X\mapsto X', Z \mapsto Z', Y\mapsto Y'\}\\
        \alpha_{X'}&:= \domain(X)\overset{identity}{\longmapsto}\domain(X')\\
        \alpha_{Z'}&:= \domain(Z)\overset{identity}{\longmapsto}\domain(Z')\\
        \alpha_{Y'}&:= \begin{cases}
                           Y \mapsto 0 & Y\text{ is even}\\
                           Y \mapsto 1 & Y\text{ is odd}
                       \end{cases}
    \end{align*}

    One can verify that this abstraction is functionally $\ltwo$ consistent. However, through the surjective mapping $\alpha_{Y'}: Y \rightarrow Y'$ the dependence of $Y'$ on $Z'$ is removed. Essentially what we have done here is summing out the effect of $Z$ over the possible values of $X$.

    Now if one is to construct the CDAG of $\model$ given by $\varphi$ we get:
    
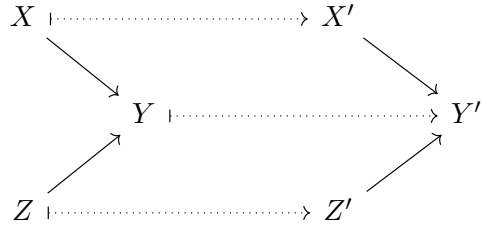
\begin{center}
        \begin{tikzcd}
            X \arrow[dr]\arrow[|->, dotted, rrr] & && X' \arrow[dr] & \\
            & Y\arrow[|->, dotted, rrr] & && Y'\\
            Z\arrow[ur]\arrow[|->, dotted, rrr]& && Z' \arrow[ur] &
        \end{tikzcd}
    \end{center}
    with all edges preserved according to the rules of the CDAG.

    However, we have shown that the dependency of $Y'$ on $Z'$ no longer exists, and therefore the edge $Z'\rightarrow Y'$ is not faithful. This illustrates that the CDAG is not necessarily compatible with surjective $\ltwo$-consistent $\alpha$-abstractions.
\end{example}
    
\subsection{Definition Constructive $\tau$ abstraction}\label{app:def-tau-omega}
    \begin{definition}
    Let $Rst(\vars,x) = \{v\in \domain(V)\vert x\text{ is the restriction of }v\text{  to }X\}$ and let $\tau: \domain(\vars_{\model})\rightarrow \clusts_{\model'}$ be given, then define $\omega_\tau(do(V_i = v_i)) = do(\clusts_i = \mathbf{c}_i)$ if $\clusts_i\in \clusts$, $\mathbf{c}_i \in \domain(\clusts_i)$ and $\tau(Rst(V_i,v_i))=Rst(\clusts_i, \mathbf{c}_i)$. \citep{beckers2019abstracting}
\end{definition}

\begin{definition}[Constructive $\tau$ abstraction {\normalfont\citep{beckers2019abstracting}}]
    Let $\model$, and $\model'$ be SCMs, then $\model'$ is a constructive $\tau$-abstraction of $\model$ if there are:

    \begin{itemize}
        \item a surjective function $\tau: \domain(\vars_{\model})\rightarrow \domain(\clusts_{\model'})$, such that there exists a partition $P=\{Z_1,\dots,Z_{n+1}\}$ of $\vars$ with $Z_1,\dots,Z_{n}$ non-empty and mappings $\tau_i:\domain(Z_{i})\rightarrow\domain(\clusts_i)$ for $i=1,\dots,n$ such that $\tau=(\tau_1,\dots,\tau_n)$.
        \item a surjective function $\tau_{\exos} : \domain(\exos_{\model}) \rightarrow \domain(\exos_{\model'})$ mapping the exogenous range of $\exos_{\model}$ to the exogenous range of $\exos_{\model'}$, compatible with $\tau$.
        \item the intervention set $\mathcal{I}_{\model'} = \omega_\tau(\mathcal{I}_{\model})$, such that all $\mathcal{I}_{\model'}$ contains all possible interventions in $\model'$.
    \end{itemize}
\end{definition}

$\tau_\exos:\domain(\exos_{\model})\rightarrow \domain(\exos_{\model'})$ is compatible with $\tau:\domain(\vars_{\model})\rightarrow\domain(\vars_{\model'})$ if for all $do(V = v)\in \mathcal{I}_{\model}$ and $\mathbf{u} \in \domain(\exos_{\model})$, $\tau(\model(\mathbf{u}, do(V=v))) = \model'(\tau_\exos(\mathbf{u}),\omega(do(V=v)))$. \citep{beckers2019abstracting}

\section{Proofs}
\subsection{Proof of lemma \ref{lem:l2-implies-l1-cons}}\label{proof-l2-implies-l1-cons}
    \begin{customlemma}{\ref{lem:l2-implies-l1-cons}}
    $\ltwo$-consistency implies $\lone$-consistency.
    \begin{proof}\normalfont
        \sloppy Let $\boldsymbol{\alpha}: \model \rightarrow \model'$ be an $\ltwo$-consistent abstraction, and $\mathbf{X},\mathbf{Y}\subseteq\vars_{\model'}$ with $\mathbf{x}\in \domain(\mathbf{X}), \mathbf{y}\in \domain(\mathbf{Y})$.

        We will show that by the null intervention $\ltwo$-consistency implies consistency over marginal and joint probabilities. We continue by showing that consistency over marginal and joint probabilities guarantee consistency for conditional distributions. Therefore, we conclude that $\ltwo$-consistency implies $\lone$ consistency. Important for this proof is surjectivity of all range maps $\alpha_{V}(v)$ to ensure the pre-image $\alpha_{V}^{-1}(v)$ exists for all $v\in\domain(V)$.

        First, by definition of $\ltwo$-consistency the following two equations hold:
        $$P(\varphi^{-1}(\mathbf{X}) = \alpha^{-1}_{\mathbf{X}}(\mathbf{x})\vert do(\emptyset)) \overset{\ltwo}{=} P(\mathbf{X} = \mathbf{x}\vert do(\emptyset)),$$
        $$P(\varphi^{-1}(\mathbf{Y}) = \alpha^{-1}_{\mathbf{Y}}(\mathbf{y}), \varphi^{-1}(\mathbf{X}) = \alpha^{-1}_{\mathbf{X}}(\mathbf{x})\vert do(\emptyset)) \overset{\ltwo}{=} P(\mathbf{Y}= \mathbf{y}, \mathbf{X} = \mathbf{x}\vert do(\emptyset)).$$
        And, by definition of the null intervention:
        $$P(\varphi^{-1}(\mathbf{X}) = \alpha^{-1}_{\mathbf{X}}(\mathbf{x})) \overset{\ltwo}{=} P(\mathbf{X} = \mathbf{x}),\text{ and}$$
        $$P(\varphi^{-1}(\mathbf{Y}) = \alpha^{-1}_{\mathbf{Y}}(\mathbf{y}), \varphi^{-1}(\mathbf{X}) = \alpha^{-1}_{\mathbf{X}}(\mathbf{x})) \overset{\ltwo}{=} P(\mathbf{Y}= \mathbf{y}, \mathbf{X} = \mathbf{x}).$$
        It follows from these equations that $\ltwo$-consistency implies:
        $$\frac{P(\varphi^{-1}(\mathbf{Y}) = \alpha^{-1}_{\mathbf{Y}}(\mathbf{y}), \varphi^{-1}(\mathbf{X}) = \alpha^{-1}_{\mathbf{X}}(\mathbf{x}))}{P(\varphi^{-1}(\mathbf{X}) = \alpha^{-1}_{\mathbf{X}}(\mathbf{x}))}\overset{\ltwo}{=}\frac{P(\mathbf{Y}= \mathbf{y}, \mathbf{X} = \mathbf{x})}{P(\mathbf{X} = \mathbf{x})}$$
        Finally, by applying Bayes' Theorem the ratios can be substituted by conditional distributions:
        $$P\left(\varphi^{-1}(\mathbf{Y}) = \alpha^{-1}_{\mathbf{Y}}(\mathbf{y})\vert \varphi^{-1}(\mathbf{X}) = \alpha^{-1}_{\mathbf{X}}(\mathbf{x})\right) \overset{\ltwo}{=} P\left(\mathbf{Y}= \mathbf{y}\vert \mathbf{X} = \mathbf{x}\right)$$
        Note that the last equation immediately implies $\lone$-consistency. Therefore, any $\ltwo$-consistent abstraction $\boldsymbol{\alpha}: \model \rightarrow \model'$ is necessarily $\lone$-consistent.

    \end{proof}
\end{customlemma}

\subsection{Proof of lemma \ref{lem:consistency-Equalities}}\label{proof-consistency-equalities}
    \begin{customlemma}{\ref{lem:consistency-Equalities}}[Bijective Range Maps Preserve Distribution (In)Equalities]
    \sloppy Let $\boldsymbol{\alpha}: \model \rightarrow \model'$ be an $\ltwo$-consistent abstraction with $\alpha_V:\domain\left(\varphi^{-1}(V)\right)\rightarrow\domain\left(V\right)$ bijective for all $V\in\vars_{\model'}$. Let $\mathbf X, \mathbf Y_1,\mathbf Y_2, \mathbf Z_1,\mathbf Z_2 \subseteq \vars_{\model}$ be partitions defined by $\varphi$, then $P(\mathbf X\:|\:do(\mathbf Y_1), \mathbf Z_1) = P(\mathbf X\:|\:do(\mathbf Y_2), \mathbf Z_2)$ if and only if $\alpha_{\mathbf{X}}[P(\mathbf X\:|\:do(\mathbf Y_1), \mathbf Z_1)] = \alpha_{\mathbf{X}}[P(\mathbf X\:|\:do(\mathbf Y_2), \mathbf Z_2)]$

    \begin{proof} \normalfont 
        Let $\boldsymbol{\alpha}: \model \rightarrow \model'$ be a $\ltwo$-consistent abstraction with $\alpha_V: \domain\left(\varphi^{-1}(V)\right)\rightarrow\domain\left(V\right)$ bijective for all $V\in\vars_{\model'}$. Let $\mathbf X, \mathbf Y_1,\mathbf Y_2, \mathbf Z_1,\mathbf Z_2 \subseteq \vars_{\model}$, such that $\mathbf X, \mathbf Y_1,\mathbf Y_2, \mathbf Z_1,\mathbf Z_2$ are pre-images of $\varphi$.
        
        This proof will consist of two parts, first we show that surjectivity of the range map gives us that 
        \begin{align*}
            \alpha_{\mathbf{X}}[P(\mathbf X\:|\:do(\mathbf Y_1), \mathbf Z_1)] &\neq \alpha_{\mathbf{X}}[P(\mathbf X\:|\:do(\mathbf Y_2), \mathbf Z_2)]\\
            &\hspace{-8pt}\text{implies}\\
            P(\mathbf X\:|\:do(\mathbf Y_1), \mathbf Z_1) &\neq P(\mathbf X\:|\:do(\mathbf Y_2), \mathbf Z_2).
        \end{align*}
        Secondly, we show that bijectivity gives us the inverse: 
        \begin{align*}
            P(\mathbf X\:|\:do(\mathbf Y_1), \mathbf Z_1) &\neq P(\mathbf X\:|\:do(\mathbf Y_2), \mathbf Z_2)\\
            &\hspace{-8pt}\text{implies} \\
            \alpha_{\mathbf{X}}[P(\mathbf X\:|\:do(\mathbf Y_1), \mathbf Z_1)] &\neq \alpha_{\mathbf{X}}[P(\mathbf X\:|\:do(\mathbf Y_2), \mathbf Z_2)],
        \end{align*} completing the proof.

        \paragraph{Part 1.}
        First consider the case where $$\alpha_{\mathbf{X}}[P(\mathbf X\:|\:do(\mathbf Y_1), \mathbf Z_1)] \neq \alpha_{\mathbf{X}}[P(\mathbf X\:|\:do(\mathbf Y_2), \mathbf Z_2)].$$ Surjectivity ensures that the distributions have pre-images in $\alpha_{\mathbf{X}}$. Given a surjective function $g$ if $g(a) \neq g(b)$ then $a\neq b$, and since pre-images exist for both distributions, these must also not be equal. So, 
        \begin{align*}
            \alpha_{\mathbf{X}}[P(\mathbf X\:|\:do(\mathbf Y_1), \mathbf Z_1)] &\neq \alpha_{\mathbf{X}}[P(\mathbf X\:|\:do(\mathbf Y_2), \mathbf Z_2)] \\
            &\hspace{-8pt}\text{implies}\\
            P(\mathbf X\:|\:do(\mathbf Y_1), \mathbf Z_1) &\neq P(\mathbf X\:|\:do(\mathbf Y_2), \mathbf Z_2).
        \end{align*}

        \paragraph{Part 2.}
        Secondly, consider the case where $$P(\mathbf X\:|\:do(\mathbf Y_1), \mathbf Z_1) \neq P(\mathbf X\:|\:do(\mathbf Y_2), \mathbf Z_2).$$ Bijectivity of $\alpha_{\mathbf{X}}$ ensures there exists an inverse to $\alpha_{\mathbf{X}}$ such that 
        \begin{align*}
            \alpha_{\mathbf{X}}^{-1}[\alpha_{\mathbf{X}}[P(\mathbf X\:|\:do(\mathbf Y_1), \mathbf Z_1)]] &= P(\mathbf X\:|\:do(\mathbf Y_1), \mathbf Z_1)\\
            &\hspace{-24pt}\text{and}\\ 
            \alpha_{\mathbf{X}}^{-1}[\alpha_{\mathbf{X}}[P(\mathbf X\:|\:do(\mathbf Y_2), \mathbf Z_2)]] &= P(\mathbf X\:|\:do(\mathbf Y_2), \mathbf Z_2).
        \end{align*} Again, consider that given a function $g$ if $g(a) \neq g(b)$ then $a\neq b$. So since 
        \begin{align*}
            P(\mathbf X\:|\:do(\mathbf Y_1), \mathbf Z_1) &\neq P(\mathbf X\:|\:do(\mathbf Y_2), \mathbf Z_2)\\
            &\hspace{-24pt}\text{it must hold that} \\
            \alpha_{\mathbf{X}}[P(\mathbf X\:|\:do(\mathbf Y_1), \mathbf Z_1)] &\neq \alpha_{\mathbf{X}}[P(\mathbf X\:|\:do(\mathbf Y_2), \mathbf Z_2)].
        \end{align*}

        \paragraph{Conclusion.}
        We have illustrated that, given a bijective $\ltwo$-consistent abstraction $\boldsymbol{\alpha}$, 
        \begin{align*}
             P(\mathbf X\:|\:do(\mathbf Y_1), \mathbf Z_1) &\neq P(\mathbf X\:|\:do(\mathbf Y_2), \mathbf Z_2)\\
             &\hspace{-12pt} \text{if and only if} \\
             \alpha_{\mathbf{X}}[P(\mathbf X\:|\:do(\mathbf Y_1), \mathbf Z_1)] &\neq \alpha_{\mathbf{X}}[P(\mathbf X\:|\:do(\mathbf Y_2), \mathbf Z_2)],
        \end{align*} and therefore also 
        \begin{align*}
            P(\mathbf X\:|\:do(\mathbf Y_1), \mathbf Z_1) &= P(\mathbf X\:|\:do(\mathbf Y_2), \mathbf Z_2)\\
            &\hspace{-12pt} \text{if and only if} \\
            \alpha_{\mathbf{X}}[P(\mathbf X\:|\:do(\mathbf Y_1), \mathbf Z_1)] &= \alpha_{\mathbf{X}}[P(\mathbf X\:|\:do(\mathbf Y_2), \mathbf Z_2)].
        \end{align*}
    \end{proof}
\end{customlemma}

\subsection{Proof of lemma \ref{lem:interventional-consistency-adjacency}}\label{proof-preservation-adjacencies}
    \begin{customlemma}{\ref{lem:interventional-consistency-adjacency}}[$\ltwo$-Consistency Uniquely Determines Adjacencies]
    Given an $\ltwo$-consistent bijective abstraction $\boldsymbol{\alpha}: \mathcal M\rightarrow \mathcal M'$, adjacencies between variables in $\mathcal M'$ are uniquely determined by the map $\varphi: \mathbf V \rightarrow \mathbf C$ and comply with Rule 1 of Def.\ref{def:CDAG} (CDAG).

    \begin{proof}\normalfont
        The proof follows from the generalized proof of Thm.\ref{lem:interventional-consistency-adjacency}. Specifically, Rule 1 of Def.\ref{def:PCDAG} (PCDAG) is equivalent to Rule 1 of Def.\ref{def:CDAG} (CDAG) for any abstraction with $\relevant = \vars_{\model}$.
    \end{proof}
\end{customlemma}
    
\subsection{Proof of lemma \ref{lem:L2-conf}}\label{proof-L2-confounding}
    \begin{customlemma}{\ref{lem:L2-conf}}[$\ltwo$-Consistency Uniquely Determines Confounding Edges]
    Given an $\ltwo$-consistent bijective abstraction $\boldsymbol{\alpha}: \mathcal M\rightarrow \mathcal M'$, confounding edges between variables in $\mathcal M'$ are uniquely determined by the map $\varphi: \mathbf V \rightarrow \mathbf C$ and comply with Rule 2 of Def.\ref{def:CDAG} (CDAG).

    \begin{proof}\normalfont
        The proof follows from the generalized proof of Thm.\ref{lem:L2-conf-2}. Specifically, Rule 2A of Def.\ref{def:PCDAG} (PCDAG) is equivalent to Rule 2 of Def.\ref{def:CDAG} (CDAG) for any abstraction with $\relevant = \vars_{\model}$, and any edges produced by rule 2B of the PCDAG require a non-empty remainder set.
    \end{proof}
\end{customlemma}
    
\subsection{Proof of theorem \ref{thm:CDAG-alpha}}\label{proof-cdag-alpha}
    \begin{customthm}{\ref{thm:CDAG-alpha}}[$\alpha$-Abstraction $\Rightarrow$ CDAG]
     \sloppy Given any bijective $\ltwo$-consistent $\alpha$-abstraction $\boldsymbol{\alpha}: \model\rightarrow\model'$ with $\relevant = \vars$,  the DAG of $\model'$ is a CDAG of the graph of $\model$.

    \begin{proof} \normalfont
        This proof follows immediately from the results of Lem.\ref{lem:interventional-consistency-adjacency} and Lem.\ref{lem:L2-conf}. Together the lemmas state that both all directed and all confounding edges of a bijective $\ltwo$-consistent abstraction with $\relevant=\vars_{\model}$ are necessarily equal to those given by the constructive rules of the CDAG.
    \end{proof}
\end{customthm}

\subsection{Proof of theorem \ref{thm:alpha-cdag}\label{proof-alpha-cdag}}
    \begin{customthm}{\ref{thm:alpha-cdag}}[CDAG $\Rightarrow$ $\alpha$-Abstraction]
    \sloppy Given a CDAG there exists an equivalent $\ltwo$-consistent $\alpha$-abstraction $\boldsymbol{\alpha}:\model\rightarrow\model'$ with $\relevant = \vars$, $\varphi$ given by the clustering, and all maps $\alpha_{\clusts}$ bijective.
    
    \begin{proof}\normalfont
         The proof follows immediately from the result of Thm.\ref{thm:PCDAG-L2-Consistent}, showing there exists an $\ltwo$-consistent $\alpha$-abstraction for each PCDAG by taking bijective range maps $\alpha_\clusts$ and the set of mechanisms $\funcs_{\model'}$ to be the composite mechanisms as implied by $\varphi$. Note that a CDAG is by deinition also a PCDAG, so the result transfers.
    \end{proof}
    \end{customthm}

\subsection{Proof of lemma \ref{lem:PCDAG-Adjancency}}\label{proof-PCDAG-adj}
    \begin{customlemma}{\ref{lem:PCDAG-Adjancency}}[PCDAGs Preserve Mediated Adjancencies]
    Let $\mathcal M$ be an SCM with a PCDAG defined by clusters $\mathbf C$ and remainder set $\remain$. Let $\mathbf C_i, \mathbf C_j \in \mathbf C, \; \mathbf C_i \neq \mathbf C_j$ and $V_i\in \mathbf C_i,\;V_j\in \mathbf C_j$, then a mediated adjacency between $V_i$ and $V_j$ exists if and only if there exists an adjacency between $\mathbf C_i$ and $\mathbf C_j$.
    
    \begin{proof}\normalfont
        It follows directly from Def.\ref{def:PCDAG}.1 (Partial Cluster DAG) that an mediated adjacency between $V_i \in \mathbf C_i$ and $V_j \in \mathbf C_j$ creates an edge between $\mathbf C_i$ and $\mathbf C_j$. It follows that, given two clusters $\mathbf C_i, \mathbf C_j \in \mathbf C$, if there exists an mediated adjacency between two variables $V_i\in \mathbf C_i, V_j \in \mathbf C_j$ in $\mathcal M$ there exists an adjacency between $\mathbf C_i$ and $\mathbf C_j$ in $\mathcal M'$.

        Furthermore, this is the only rule creating a directed edge, and therefore solely defines adjacencies in a PCDAG. It follows that, given two clusters $\mathbf C_i, \mathbf C_j \in \mathbf C$, if there does not exists some combination $V_i\in \mathbf C_i, V_j \in \mathbf C_j$ in $\mathcal M$ with an mediated adjacency between $V_i$ and $V_j$ then there is no adjacency between $\mathbf C_i$ and $\mathbf C_j$ in $\mathcal M'$.

        Thus showing that for each mediated adjacency in $\model$ there exists a corresponding adjacency in $\model'$, and for all adjacencies in $\model'$ there exists at least one corresponding mediated adjacency in $\model$.
    \end{proof}
\end{customlemma}

\subsection{Proof of lemma \ref{lem:PCDAG-Directed-Paths}}\label{proof-PCDAG-dir-paths}
    \begin{customlemma}{\ref{lem:PCDAG-Directed-Paths}}[PCDAGs Preserve Directed Paths]
    Let $\mathcal M$ be an SCM with a PCDAG defined by clusters $\mathbf C$ and remainder set $\remain$. Let $\mathbf C_i, \mathbf C_j \in \mathbf C, \; \mathbf C_i \neq \mathbf C_j$ and $V_i\in \mathbf C_i,\;V_j\in \mathbf C_j$, then a directed path $\mathbf C_i \rightarrow \dots \rightarrow \mathbf C_j$ exists if there exists a directed path $V_i \rightarrow \dots \rightarrow V_j$.
    
    \begin{proof}\normalfont
        First, by Def.\ref{def:PCDAG}.1 (Partial Cluster DAG) adjacencies in the PCDAG have the same direction as the mediated adjacencies in the original DAG. Secondly, by Lem.\ref{lem:PCDAG-Adjancency} adjacencies in the PCDAG are consistent with mediated adjacencies in $\mathcal M$. Note that adjacency are directed paths of length 1, so if follows immediately that such directed paths are preserved.

        Now consider directed paths of length greater than 1. Let $V_i\in \mathbf C_i,\;V_j\in \mathbf C_j$ with $\mathbf C_i, \mathbf C_j \in \mathbf C$ and let there be a directed path $V_i \rightarrow \dots \rightarrow V_j$. Any directed path can be split up into a sequence of mediated adjacencies. Take $V_i$ and the first vertex $V_x$ along the path such that $V_x \notin \mathbf R$, then either (i) there exists an edge $V_i \rightarrow V_x$ or (ii) there exist a path $V_i \rightarrow \dots \rightarrow V_x$ such that any intermediate vertices are in $\mathbf R$. Therefore, there exists a mediated adjacency between $V_i$ and $V_x$, by Lem.\ref{lem:PCDAG-Adjancency} this implies $\mathbf C_i \rightarrow\mathbf C_x, \; V_x\in \mathbf C_x$ if $\mathbf C_x \neq \mathbf C_i$. By repeated application of this process starting from $V_x$, until we reach $V_j$, all generated edges between clusters can be composed into the directed path $\mathbf C_i \rightarrow\dots \rightarrow\mathbf C_j$.
    \end{proof}
\end{customlemma}

\subsection{Proof of theorem \ref{thm:PCDAG-L2-Consistent}}\label{proof-PCDAG-Consistency}
    \begin{customthm}{\ref{thm:PCDAG-L2-Consistent}}[$\ltwo$-consistency of PCDAGs]
    Given an SCM $\model$ and a PCDAG $\model'$ of $\model$, there exists a set of mechanisms $\mathcal F_{\model'}$ such that $\model'$ is $\ltwo$-consistent with $\model$.
    \begin{proof} \normalfont
        For this proof we will show that a given two models $\model, \model'$ where $\model'$ is PCDAG of $\model$, there exists a configuration of $\model'$ such that $\model'$ is $\ltwo$ consistent with $\model$. We will define the functions for $\model'$ according to the clustering, and show that these functions correspond to the structure of the PCDAG. After we show that cluster variables together with these functions give us $\ltwo$ consistency.
    
        First, let $\model : \langle \exos_{\vars}, \vars, \funcs_{\vars}, \dists{\vars}\rangle$ and $\model' : \langle \exos_{\clusts}, \clusts, \funcs_{\clusts}, \dists{\clusts}\rangle$ be two SCMs such that the DAG of $\model'$ is a PCDAG of $\model$ and $\clusts$ the corresponding clusters.

        We choose $\exos_{\clusts} = \exos_{\vars}$ and $\dists{\clusts}=\dists{\vars}$, and let the variables $\clusts_i\in\clusts$ be given by the clustering such that 
        $$
        \clusts_i = \begin{pmatrix}
            V^i_1\\
            \vdots\\
            V^i_n
        \end{pmatrix},
        $$ 
        with $n$ as the number of variables in cluster $\clusts_i$ and $V^i_j\in \clusts_i$ as the $j$-th variable in the cluster $\clusts_i$. Now construct the functions $f_{\clusts_i}\in \funcs_{\clusts}$ again following the clustering:
        $$
        f_{\clusts_i} = \begin{pmatrix}
            f_{V^i_1}(Pa(V^i_1), \exos_{V^i_1})\\
            \vdots\\
            f_{V^i_n}(Pa(V^i_n), \exos_{V^i_n})
        \end{pmatrix},
        $$
        with $f_{V^i_j} \in \funcs_{\vars}$ the function for $V^i_j$ in $\model$, and $\exos_{V^i_j}$ the exogenous parents of $V^i_j$.

        This definition of $f_{\clusts_i}$ leaves two potential issues:
        \begin{enumerate}
            \item There can exist a function $f_{V^i_j}$ part of $f_{\clusts_i}$ which relies on a variable mapped to $\remain$. By Def.\ref{def:PCDAG} of the PCDAG there exists no edges to or from $\remain$, therefore such reliance must somehow be removed.

            \item There can exist a function $f_{V^i_j}$ part of $f_{\clusts_i}$ which relies on a variable $V^i_k$, which is also mapped to $\clusts_i$. Acyclicity requires there exist no edge $\clusts_i\rightarrow \clusts_i$, therefore such reliance must also be removed.
        \end{enumerate}

        \sloppy First we describe how we can rewrite functions in the case of 1.: take $f_{V^i_j}(Pa(V^i_j), \exos_{V^i_j})$ part of $f_{\clusts_i}$ with $Pa(V^i_j)\cap \remain \neq \emptyset$. Let $\mathbf P = Pa(V^i_j)\cap \remain$ be the set of parents of $V^i_j$ that are mapped to $\remain$. All variables $P \in \mathbf P$ can be substituted with their functions instead. These functions will in turn depend on a set of parents and exogenous parents. By acyclicity recursively applying this technique all functions will have $Pa(P) \cap \remain = \emptyset$ in a finite number of applications. When $Pa(P) \cap \remain = \emptyset$ all parents of $P$ are part of some cluster, or there are no endogenous parents left. Note that this gets reliance on exogenous parents of all $P$s encountered.

        For case 2. we follow a similar approach: take $f_{V^i_j}(Pa(V^i_j), \exos_{V^i_j})$ part of $f_{\clusts_i}$ with some subset $\mathbf P \subseteq Pa(V^i_j)\cap \clusts_i \neq \emptyset$. Applying the same substitution technique as before gives a function that only relies on variables outside of the cluster $\clusts_i$.

        These functions agree with the structure given by Def.\ref{def:PCDAG} of the PCDAG as causal edges $\clusts_i \rightarrow\clusts_j$ exists if and only if there exists some $V_i\in \clusts_i$ and $V_j\in \clusts_j$ such that there is a mediated adjacency $V_i\rightsquigarrow V_j$. Furthermore, recall that $\clusts_i \confarrow \clusts_j$ is a shorthand for $\clusts_i \leftarrow U \rightarrow \clusts_j$. Given the definition of the functions $\funcs_{\clusts}$ such confounding edges occur either (i) if there exists $V_i\in \clusts_i$ and $V_j\in \clusts_j$ such that $V_i\leftarrow U \rightarrow V_j$, or (ii) $V_i\in \clusts_i$, $V_j\in \clusts_j$, and $Q\in\remain$ such that 
        \begin{center}\begin{tikzcd}[cramped, sep=small]
            V_i & Q\arrow[l]\arrow[r] & V_j\\
            & U\arrow[u]&
        \end{tikzcd}.
        \end{center}
        Note that these cases exactly correspond to Def.\ref{def:PCDAG} (Partial Cluster DAG). This shows that the definition of $\funcs_{\clusts}$ is compatible with the structure defined by the PCDAG.

        This leaves to show that such definitions give rise to $\ltwo$-consistency. For this, first note that by clustering there is a bijection between the ranges of a cluster $\clusts_i\in \clusts$ and the variables that make up the cluster $\mathbf V_i \subseteq \vars$. Additionally, the function $f_{\clusts_i}$ is simply a combination of the functions of $\mathbf V_i$. So since $\exos_{\clusts} = \exos_{\vars}$ and $\dists{\clusts} = \dists{\vars}$ it follows immediately that $$P_\clusts(\clusts_i\:|\:do(\clusts_j), \clusts_k) = P_\vars(\vars_i\:|\:do(\vars_j), \vars_k).$$ Therefore, it is immediate that there exists a model $\model'$ which is a compatible PCDAG of $\model$ that is $\ltwo$-consistent with $\model$.
    \end{proof}
\end{customthm}

\subsection{Proof of lemma \ref{lem:interventional-consistency-adjacency-2}}\label{proof-preservation-adjacencies-2}
    \begin{customlemma}{\ref{lem:interventional-consistency-adjacency-2}}[$\ltwo$-Consistency Uniquely Determines Adjacencies]
    Given a bijective $\ltwo$-consistent abstraction $\boldsymbol{\alpha}: \mathcal M\rightarrow \mathcal M'$, adjacencies between variables in $\mathcal M'$ are uniquely determined by the map $\varphi: \mathbf V \rightarrow \mathbf C$ and comply with Rule 1 of Def.\ref{def:PCDAG} (PCDAG).

    \begin{proof}\normalfont
        We will show that adjacencies in an abstracted model is uniquely determined by the map of the endogenous variables $\varphi: \mathbf V \rightarrow \mathbf C$. To get to this, we show that adjacencies $\clusts_i \rightarrow \clusts_j$ for any $\clusts_i, \clusts_j \in \clusts$ exists if, and only if, there exist an mediated adjacency between the corresponding pre-images of $\varphi$ .
        
        Let $\mathbf C$ be a (partial) clustering of $\mathbf V$. Given two variables $V_i, V_j \in \mathbf V$ such that $V_i \in \mathbf C_i$, $V_j \in \mathbf C_j$ and $\mathbf{C}_i,\mathbf{C}_j \in \mathbf{C}$, with $\; \mathbf{C}_i\neq \mathbf{C}_j$. 
        
        First, consider the case where there exists an mediated adjacency $V_i \rightsquigarrow V_j$. By definition $P(V_j) \neq P(V_j\:|\: do(V_i))$, and by $\ltwo$-consistency $P(\mathbf C_j) \neq P(\mathbf C_j\:|\: do(\mathbf C_i))$, so there must exist a path $\mathbf C_i \rightarrow\mathbf C_j$. 
        Furthermore, by definition of the mediated adjacency there does not exist some $\mathbf Z \in \clusts$ that blocks the effect of $do(V_i)$ such that $$P(V_j \:|\:do(V_i),\:do(\varphi^{-1}(\mathbf Z))) = P(V_j \:|\:do(\varphi^{-1}(\mathbf Z))).$$ Lem.\ref{lem:consistency-Equalities} illustrates that 
        \begin{align*}
            P(V_j \:|\:do(V_i),\:do(\varphi^{-1}(\mathbf Z))) &= P(V_j \:|\:do(\varphi^{-1}(\mathbf Z)))\\
            &\hspace{-24pt}\text{implies} \\
            P(\clusts_j \:|\:do(\clusts_i),\:do(\mathbf Z)) &= P(\clusts_j \:|\:do(\mathbf Z)).
        \end{align*} In other words, there exists no intervention in $\model'$ that blocks the path $\clusts_i\rightarrow \clusts_j$, and since all interventions are allowed, it must be that $\clusts_i\rightarrow \clusts_j$ is an adjacency.

        Second, consider the inverse: there exists no mediated adjacency $V_i \rightsquigarrow V_j$. Then we have that there exists some $\mathbf Z\subseteq \mathbf C$, $V_i\notin \mathbf Z$ such that $$P(V_j\:|\:do(V_i),\:do(\varphi^{-1}(\mathbf Z)) = P(V_j\:|\:do(\varphi^{-1}(\mathbf Z))).$$ Interventional consistency requires that $$P(\mathbf C_j\:|\:do(\mathbf C_i),\:do(\mathbf Z)) = P(\mathbf C_j\:|\:do(\mathbf Z)).$$ So, if there exists a path $V_i \rightarrow V_j$ that can be blocked by an intervention $do(\varphi^{-1}(\mathbf Z))$ with $\mathbf Z\subseteq \mathbf C$, $\mathbf Z\neq V_j$, then the path between $\mathbf C_i \rightarrow \mathbf C_j$ can also be blocked by $do(\mathbf Z)$. Consequently, we have that if there does not exists an mediated adjacency $V_i\rightsquigarrow V_j$ then there is no adjacency $\mathbf C_i \rightarrow \mathbf C_j$.

        Therefore, for any given abstraction $\boldsymbol{\alpha}: \model \rightarrow \model'$ with $\varphi:\vars \rightarrow \mathbf C$ adjacencies in $\model'$ are uniquely given by interventional consistency of $\boldsymbol{\alpha}$. Moreover, we note that that the preservation of mediated adjacencies exactly complies to rule 1. of Def.\ref{def:PCDAG} (PCDAG).
    \end{proof}
\end{customlemma}

\subsection{Proof of lemma \ref{lem:L2-conf-2}}\label{proof-L2-confounding-2}
    \begin{customlemma}{\ref{lem:L2-conf-2}}[$\ltwo$-Consistency Uniquely Determines Confounding Edges]
    Given a bijective $\ltwo$-consistent abstraction $\boldsymbol{\alpha}: \mathcal M\rightarrow \mathcal M'$, confounding edges between variables in $\mathcal M'$ are uniquely determined by the map $\varphi: \mathbf V \rightarrow \mathbf C$ and comply with Rule 2A and 2B of Def.\ref{def:PCDAG} (PCDAG).

    \begin{proof}\normalfont
        The proof for this Lemma is split into two parts: first we show that $\ltwo$ distributions can be used to identify confounding edges between non adjacent variables. Secondly, we show the same is possible for adjacent variables, but the proof requires a different approach.

        \paragraph{Non Adjacent Confounding Edges.}
        For this part we will show that given two variables $X, Y \in \vars$ of a model $\model$ with $X$ and $Y$ not adjacent there exists some $\mathbf  Z \subseteq \vars\setminus\{X,Y\}$ such that $$P(Y\:|\:X, do(\mathbf Z)) = P(Y\:|\:do(\mathbf Z))$$ if and only if there does not exist a confounding edge $X\confarrow Y$. First we show that non existence of a confounding edge $X \confarrow Y$ implies $$P(Y\:|\:X, do(\mathbf Z)) = P(Y\:|\:do(\mathbf Z)).$$ After, we show that $X \confarrow Y$ implies $$P(Y\:|\:X, do(\mathbf Z)) \neq P(Y\:|\:do(\mathbf Z)),$$ completing this part of the proof.
        
        First, assume $X$ and $Y$ do not share a confounding edge $X\confarrow Y$. Non-adjacency gives that there exists some $\mathbf Z \subset \vars\setminus \{X,Y\}$ blocking all paths between $X$ and $Y$ that have at least one intermediate vertex in $\vars$. A path without a vertex in $\vars$ would imply $X\confarrow Y$, which by assumption do not exist. So, there exists some $\mathbf Z \subset \vars\setminus \{X,Y\}$ such that $Y\ind X \:|\:do(\mathbf Z)$. Or, in terms of distributions: $$P(Y\:|\:X, do(\mathbf Z)) = P(Y\:|\:do(\mathbf Z))$$ for all non-adjacent $X,Y$ if there does not exist a confounding edge $X\confarrow Y$. 

        Secondly, we show the inverse: assume there exists a confounding edge $X\confarrow Y$. By assumption of $X\confarrow Y$ there exists a path between $X$ and $Y$ that cannot be blocked by any intervention. So there cannot exist a $\mathbf Z \in \vars \setminus \{X,Y\}$ such that $Y\ind X \:|\:do(\mathbf Z)$. This is expressed in terms of distributions as $$P(Y\:|\:X, do(\mathbf Z)) \neq P(Y\:|\:do(\mathbf Z))$$ for all $\mathbf Z$, if there is a $X\confarrow Y$.
        
        We conclude that all confounding edges between two non-adjacent variables $X$ and $Y$ can be identified using $\ltwo$ distributions.

        \paragraph{Adjacent Confounding Edges.}
        Whereas one can identify confounding effects by blocking causal paths when considering non-adjacent variables, this is not possible in the case where $X\rightarrow Y$. The following part illustrates that $\ltwo$ distributions can still be used to identify this confounding effect through inequalities.

        Given some $X,Y\in \vars$ of $\model$, such that $X\rightarrow Y$. we will show that the absence of a confounder $X\confarrow Y$ implies $$P(Y\:|\:X,do(\mathbf Z)) = P(Y\:|\:do(X),do(\mathbf Z)).$$ Secondly, we show that the presence of a confounder $X\confarrow Y$ implies $$P(Y\:|\:X,do(\mathbf Z))\neq P(Y\:|\:do(X),do(\mathbf Z)).$$

        First, recall that any indirect path between $X$ and $Y$ can be blocked by some $\mathbf Z \in \vars \setminus\{X,Y\}$. It follows that given such a $\mathbf Z$ we have $$P(Y\:|\:X,do(\mathbf Z)) = P(Y\:|\:do(X),do(\mathbf Z)).$$

        Now let's consider the following diagram:
        \begin{center}
            \begin{tikzcd}
                X\arrow[r]\arrow[r, bend left, dashed, <->, "U"]& Y.
            \end{tikzcd}
        \end{center}
        Recall that $X\confarrow Y$ is an shorthand for writing $X\leftarrow U \rightarrow Y$. Note also that an arrow $X\rightarrow Y$ implies $P(Y)\neq P(Y\:|\:X)$ and $P(X)\neq P(X\:|\:Y)$, assuming non-canceling paths. Now consider what happens when conditioning on $X$ by taking $P(Y\:|\:X)$: there exists an arrow $U\rightarrow X$, so $P(U\:|\:X)\neq P(U)$.

        \sloppy Alternatively, consider $P(Y\:|\:do(X))$ would break the edge $U\rightarrow X$, so $P(U)=P(U\:|\:do(X))$. Furthermore the distributions $P(Y\:|\:X)$ and $P(Y\:|\:do(X))$ are determined by a function $f_Y(U, U_Y, X)$ with $U_Y$ some optional exogenous variable affecting $Y$. Therefore, given $X$ and a distribution $P(U_Y)$ the generated distributions depend only on the distribution $P(U)$, and since  $$P(U\:|\:do(X))=P(U)\neq P(U\:|\:X)$$ it follows that $$P(Y\:|\:X)\neq P(Y\:|\:do(X)).$$ In the case the model has more variables than $X$ and $Y$ any other path can be blocked by some $\mathbf Z$, similar to the first part.

        In conclusion, given two adjacent variables $X,Y$ such that $X\rightarrow Y$, there exists a confounding arrow $X\confarrow Y$ if, and only if, $$P(Y\:|\:X,do(\mathbf Z))\neq P(Y\:|\:do(X),do(\mathbf Z)).$$

        \paragraph{Preservation of Confounding Edges.} Given a bijective $\ltwo$-consistent $\alpha$-abstraction $\boldsymbol{\alpha}: \model \rightarrow \model'$, Lem.\ref{lem:consistency-Equalities} states that all (in)equalities among distributions in $\model'$ have a corresponding (in)equality in $\model$. As confounding edges can be determined uniquely by these inequalities, if $\relevant=\vars_{\model}$ then a confounding edge $\vars_i\rightarrow \vars_j$ in $\model$ implies a corresponding confounding edge $\clusts_i \confarrow \clusts_j$ in $\model'$, in accordance to rule 2A of Def.\ref{def:PCDAG} (PCDAG).

        \paragraph{Creation of Confounding edges.}
        Additionally, note that if $\relevant \subset \vars_{\model}$ such that the remainder set in non-empty: $\remain\neq\emptyset$ all $Z\in\remain$ are not considered in the preservation of (in)equalities of Lem.\ref{lem:consistency-Equalities}, and therefore can essentially be considered exogenous variables. This implies that if there exists two clusters $\clusts_i,\clusts_j\in\clusts_{\model'}$ some $Z\in\remain$ and $\vars_i\in\clusts_i$, $\vars_j\in\clusts_j$, such that $\vars_i \;\reflectbox{$\rightsquigarrow$}\; Z \rightsquigarrow \vars_j$ then there must exists a confounding arrow $\clusts_i\confarrow\clusts_j$, in accordance to rule 2B of Def.\ref{def:PCDAG}.

        Finally, as all other distributional (in)equalities are preserved over the abstraction, and all confounding edges can be identified by these (in)equalities there cannot be any other confounding edges in $\model'$. Therefore, the confounding edges of a bijective $\ltwo$-consistent abstractions are exactly described by the constructive rules of the PCDAG.
        
    \end{proof}
\end{customlemma}

\subsection{Proof of theorem \ref{thm:PCDAG-alpha}}\label{proof-pcdag-alpha}
    \begin{customthm}{\ref{thm:PCDAG-alpha}}[PCDAGs describe all bijective $\ltwo$-consistent $\alpha$-abstractions]
    \sloppy Given a bijective $\ltwo$-consistent $\alpha$-abstraction $\boldsymbol{\alpha}: \model\rightarrow\model'$, $\model'$ is a permissible Partial Cluster DAG of $\model$.

    \begin{proof} \normalfont
        This proof follows immediately from the results of Lem.\ref{lem:interventional-consistency-adjacency-2} and Lem.\ref{lem:L2-conf-2}. Together the lemmas state that both all directed and all confounding edges of a bijective $\ltwo$-consistent abstraction are necessarily equal to those given by the constructive rules of the PCDAG.
    \end{proof}
\end{customthm}

\subsection{Proof of theorem \ref{thm:alpha-PCDAG}\label{proof-alpha-PCDAG}}
    \begin{customthm}{\ref{thm:PCDAG-alpha}}[PCDAG $\Rightarrow$ $\alpha$-Abstraction]
     \sloppy Given a PCDAG there exists an equivalent $\ltwo$-consistent $\alpha$-abstraction $\boldsymbol{\alpha}:\model\rightarrow\model'$ with $\relevant \subseteq \vars$, $\varphi$ given by the clustering, and all maps $\alpha_{\clusts}$ bijective.

     \begin{proof}\normalfont
         The proof follows immediately from the result of Thm.\ref{thm:PCDAG-L2-Consistent}, showing there exists an $\ltwo$-consistent $\alpha$-abstraction for each PCDAG by taking bijective range maps $\alpha_\clusts$ and the set of mechanisms $\funcs_{\model'}$ to be the composite mechanisms as implied by $\varphi$.
     \end{proof}
    \end{customthm}

\subsection{Proof of Corollary \ref{cor:alpha-constructive-tau}}\label{proof-alpha-constructive-tau}
    \begin{customcorollary}{\ref{cor:alpha-constructive-tau}}[Equivalence $\alpha$-abstraction and Constructive $\tau$-abstraction]
    \sloppy The $\alpha$-abstraction is equivalent to the constructive $\tau$-abstraction, if for all settings $\mathbf{v}\in\domain(\vars)$ there exists a $\mathbf{u}\in\domain(\exos)$ giving rise to $\mathbf{v}$.
    
    \begin{proof} \normalfont
        In this proof we will first outline the similarities of the two abstractions, by unpacking the definitions. After that we will show that the frameworks do not as neatly align on mapping the exogenous variables. In short, the problem arises from the fact that the $\alpha$-abstraction is not defined over SCMs but Finite Graphical Models. 

        \paragraph{Similarities $\alpha$-abstraction and constructive $\tau$-abstraction.} 
        This part will be rather straightforward, as we will see that all maps that make up the abstractions are equivalent by definition.

        Let $\boldsymbol{\alpha}: \model \rightarrow \model'$ be an $\alpha$-abstraction following definition \ref{def:alpha-abs}.
        \begin{enumerate}
            \item First note that the map $\varphi:\mathbf R \rightarrow \vars_{\model'}$ of $\boldsymbol{\alpha}$ is surjective. It follows immediately that $\varphi$ induces a partition $\mathbf P$ over $\relevant$. Consequently, $\mathbf P$ together with the set of irrelevant variables forms a partition over $\vars_\model$, as required by the constructive $\tau$-abstraction.
            \item Secondly, given an abstract variable $X'\in\vars_{\model'}$ and its pre-image $\mathbf X = \varphi^{-1}(X')$ the range mapping $\domain(\mathbf X)\rightarrow\domain(X')$ is surjective by definition. This is immediately equivalent to the subjectivity requirement imposed on range mappings in the constructive $\tau$-abstraction.
            \item thirdly, the constructive $\tau$ abstraction requires that the set of allowed interventions at the high level is equal to the set of all high level interventions. In other words, there must exist a surjective map from the set of low level interventions to the set of high level interventions. Given an $\alpha$-abstraction $\boldsymbol{\alpha}: \model \rightarrow \model'$ the intervention map is implicitly given by $P(\varphi^{-1}(Y)\:|\:do(\varphi^{-1}(X))) \rightarrow P(Y\:|\:do(X))$, for all $X, Y \in \vars_{\model'}$. It follows from surjectivity of $\varphi$ and the range maps $\domain(\varphi^{-1}(X))\rightarrow \domain(X)$ that the intervention map is also surjective.
        \end{enumerate}

        \paragraph{Finite Graphical Model as a minimal SCM.}
        Finally, a constructive $\tau$-abstraction is by definition a $\tau$-abstraction, which requires that there exists a surjective map from range of the exogenous variables of the low level model to that of the high level model. Here is the problem, as the exogenous variables are not explicitly defined for the Finite Graphical Model. However, one can construct a minimal range for the exogenous variables required to generate the distributions defined by the finite graphical model:

        Let $\model : \langle \vars, \exos, \funcs, \dists{}\rangle$ be an SCM. Given a setting of the exogenous variables $\textbf{u}$ there exists exactly one compatible setting of the endogenous variables $\textbf{v}$, by the fact that all $f\in \funcs$ are deterministic. This means there is a surjective function from the range of $\exos$ to the range of $\vars$. Consequently, we can define a partition of equivalence classes on the range of $\exos$, such that for each equivalence class all settings of $\exos$ in that class will map to the same setting of $\vars$. Therefore, the minimal range of $\exos$ is such that there exists exactly one value for each equivalence class. It follows immediately that there exists a bijective map from the minimal range of $\exos$ to the range of $\vars$.

        let $\boldsymbol{\alpha}: \model \rightarrow \model'$ be an alpha abstraction, by definition we have $\varphi: \domain(\vars_\model)\rightarrow \domain(\vars_{\model'})$. Let $\exos^*_{\model}$ and $\exos^*_{\model'}$ be minimal exogenous variables for $\model$ and $\model'$ respectively. There is a bijection between $\vars_\model$ and $\exos^*_{\model}$ and $\varphi$ surjective, therefore there exists a $\varphi_U: \exos^*_{\model} \rightarrow \exos^*_{\model'}$ that is surjective.

        The $\tau$-abstraction requires a surjective map $\tau_U: U_\model \rightarrow U_{\model'}$, which is not necessary for the $\alpha$-abstraction. However, we have shown that the $\alpha$-abstraction does guarantee a surjective  map $\exos^*_{\model}\rightarrow\exos^*_{\model'}$. Diagrammatically this can be described as:
        
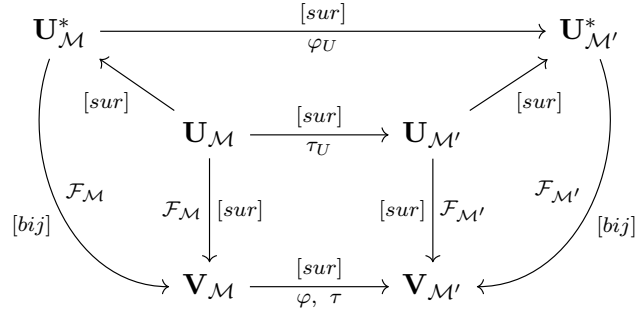
\begin{center}
            \begin{tikzcd}
                \exos^*_{\model} \arrow[rrrr, "{[sur]}", "{\varphi_U}"'] \arrow[dddr, "{[bij]}"', "{\funcs_{\model}}", out=-110, in = -180]& & & & \exos^*_{\model'}\arrow[dddl, "{[bij]}", "{\funcs_{\model'}}"', out=-70, in=0]\\
                & \exos_{\model} \arrow[ul, "{[sur]}"] \arrow[rr, "{[sur]}", "{\tau_U}"'] \arrow[dd, "{[sur]}", "{\funcs_{\model}}"']& & \exos_{\model'} \arrow[ur, "{[sur]}"'] \arrow[dd, "{[sur]}"', "{\funcs_{\model'}}"] & \\
                & & & & \\
                & \vars_{\model} \arrow[rr, "{[sur]}", "{\varphi,\;\; \tau}"'] & & \vars_{\model'}& 
            \end{tikzcd}
        \end{center}
        The inner square represents the constructive $\tau$-abstraction, and the outer square represents the $\alpha$-abstraction.

        \paragraph{Conclusion.}
        We have illustrated the equivalence of the $\alpha$-abstraction and the constructive $\tau$-abstraction. With the one caveat being that the constructive $\tau$-abstraction is more restrictive on the mapping of exogenous variables.
    \end{proof}
\end{customcorollary}

\end{document}